\documentclass[10pt,twocolumn,letterpaper]{article}

\usepackage[pagenumbers]{cvpr} 

\usepackage{algpseudocode}  
\usepackage{bm}  
\usepackage{booktabs}  
\usepackage{float} 
\usepackage{gensymb}   
\usepackage{makecell, rotating}  
\usepackage{multirow}
\usepackage{stmaryrd}  

\renewcommand{\theadfont}{\normalsize}

\newcommand{\RR}{\mathbb{R}}
\DeclareMathOperator{\diff}{Diff}
\newcommand{\SE}{\mathrm{SE}}

\DeclareMathOperator*{\argmin}{arg\,min}
\DeclareMathOperator*{\argmax}{arg\,max}
\newcommand{\matr}[1]{\bm{#1}}
\newcommand{\vect}[1]{\bm{#1}}
\newcommand{\notset}[1]{\mathcal{#1}}

\newcommand{\LB}{\mathrm{LB}}
\newcommand{\UB}{\mathrm{UB}}

\newcommand{\abs}[1]{\left|#1\right|}
\newcommand{\norm}[1]{\left\|#1\right\|}
\newcommand{\set}[1]{\left\lbrace#1\right\rbrace}
\newcommand{\family}[2]{\ensuremath{\left(#1\right)_{#2}}}

\newcommand{\card}[1]{\ensuremath{\text{Card}\left(#1\right)}}
\newcommand{\tr}[1]{#1^{\top}}

\definecolor{cvprblue}{rgb}{0.21,0.49,0.74}
\usepackage[pagebackref,breaklinks,colorlinks,allcolors=cvprblue]{hyperref}

\def\paperID{}
\def\confName{CVPR}
\def\confYear{2026}

\title{Lipschitz Optimization for Formal Verification of Homographies}

\author{Jean-Guillaume Durand\\
\textit{Joby Aviation}\\
{\tt\small \href{mailto:jg.durand@jobyaviation.com}{jg.durand@jobyaviation.com}}
\and
Panagiotis Kouvaros\\
\textit{Safe Intelligence}\\
{\tt\small \href{mailto:panagiotis@safeintelligence.ai}{panagiotis@safeintelligence.ai}}
\and
Maxime Gariel\\
\textit{Joby Aviation}\\
{\tt\small \href{mailto:maxime.gariel@jobyaviation.com}{maxime.gariel@jobyaviation.com}}
\and
Alessio Lomuscio\\
\textit{Safe Intelligence}\\
{\tt\small \href{mailto:alessio@safeintelligence.ai}{alessio@safeintelligence.ai}}
}

\begin{document}

\twocolumn[{
	\renewcommand\twocolumn[1][]{#1}

	\maketitle


	\includegraphics[width=\linewidth]{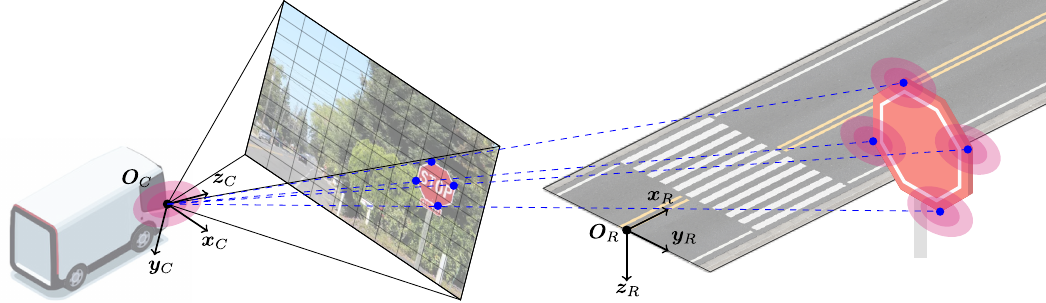}
	\captionof{figure}{\textbf{Example application: autonomous driving.}
		The scene geometry defines the coordinate transform $\bm{T}_C^R \in \SE(3)$ from road frame $(R)$ to camera frame $(C)$. Assuming a planar traffic sign and a pinhole camera model, the stop sign features (blue points) are projected from the 3D world to the 2D image plane through projective geometry. Such a plane-to-plane mapping is a homography. We then study the robustness of traffic sign recognition to uncertainty or perturbations on the camera pose and stop sign localization (magenta ellipsoids). The formal verification of such homographies can support the certification of safety-critical applications in robotics.
		\vspace{1em}
	}
	\label{fig:geometry}
}]

\begin{abstract}
The adoption of vision neural networks in regulated industries requires formal robustness guarantees, especially in safety-critical domains such as healthcare, autonomous vehicles, and aerospace. However, current approaches are confined to incomplete statistical verification or robustness to $\ell_p$-norm and affine transforms, which cover only a narrow subset of perturbations to the image formation process. In particular, robustness to camera motion remains an open problem despite being key to deploy many vision applications. We present a formal verification approach that targets robustness against 3D motion perturbations of the capturing camera. We first establish a closed-form mapping from camera pose to pixel values. By analyzing the continuity properties of the resulting homographies, we show that recent work on Lipschitz optimization and piecewise continuity can be extended to derive tight linear bounds on perturbed pixel values. Our approach applies to scenes with predominantly planar structure, such as ground planes in augmented reality, road markings and traffic signs in autonomous driving, or planar workspaces in robotic manipulation. This enables the first formal verification of projective geometry transforms, without complex simulation, surrogate networks, or explicit image-formation models. We validate our implementation and show up to 89\% speedup and 7\% tighter bounds over prior work. We then evaluate our method on the VNN-COMP benchmark and reveal systematic weaknesses to projective perturbations. Finally, we demonstrate a real-world case study on a safety-critical runway classifier, highlighting practical vulnerabilities to camera motion, and addressing a key challenge in the certification of learned models. Data and code are publicly available at \url{https://github.com/jeangud/homography-verification}.
\end{abstract}

\section{Introduction}
The deployment of deep neural networks in safety-critical computer
vision systems is rapidly accelerating across diverse applications, from augmented reality and medical imaging, to autonomous vehicles and aerospace (\cref{fig:geometry}). Deep learning has proven a key enabler, significantly improving system performance and generalization~\cite{lecun2015deep,balduzzi2021neural,ducoffe2023lard}. However, these models exhibit instabilities and non-linear behaviors~\cite{szegedy2014intriguing, goodfellow2014explaining}, complicating robustness analysis and ultimately their regulatory compliance. Notwithstanding, for these systems to be trusted and certified in regulated domains, they require formal guarantees demonstrated under foreseeable operational scenarios~\cite{wasson2024deobfuscating}. Such robustness evidence is often produced by formal methods, which provide strong mathematical guarantees on system outputs.\\

\begin{figure}
	\centering
	\includegraphics[width=\linewidth]{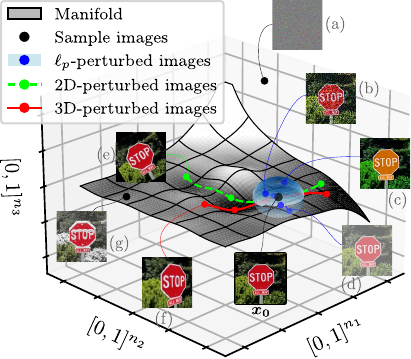}
	\caption{\textbf{The space of image perturbations (notional).}
		Most vectors in $[0,1]^{h \times w}$ are noise (a). The subset of natural images forms a complex manifold of lower dimension (grey surface), on which lies our original image $\vect{x_0}$. The ball $B_p(\vect{x_0}, \varepsilon)$ of $\ell_p$-norm perturbations (blue) can capture noise (b), hue (c) and brightness (d) variations for example. Beyond $B_p(\vect{x_0}, \varepsilon)$ and along constrained walks on the manifold, we may follow affine 2D perturbations (green) such as translation or rotation (e). We also encounter 3D perturbations (red): the same scene captured from a different perspective (f). Further on the manifold, lie more complex semantic perturbations like lens flare or snow (g). More natural images (e.g. different signs, cats) form other subsets of this manifold.
	}
	\label{fig:manifold}
\end{figure}

To date, formal verification for neural networks (VNN) has largely focused on robustness against $\ell_p$-bounded perturbations~\cite{brix2023vnncomp, li2023sok}, which act directly on the pixel space (e.g. contrast, color). This approach may not model some scenarios where perturbations originate in the physical space of the scene (e.g. viewpoint, shadows), before impacting image formation. While a sufficiently large $\ell_p$-ball might contain the resulting images, it fails to capture the essential constraints defining the transformation. This approach also over-approximates the true manifold of perturbed images (\cref{fig:manifold}), leading to uninformative loose bounds that are insufficient to prove network robustness~\cite{engstrom2019exploring,mangal2023certifying}.\\

Consequently, verifying robotics applications requires handling a comprehensive range of perturbations within the Operational Design Domain (ODD), including 3D scene transformations. In this context, homographies emerge as a crucial class of transforms that capture perspective changes: they describe the inherent projective mapping between two different views of a planar surface~\cite{hartley2003multiple}. This plane-to-plane mapping is ubiquitous in computer vision and is central to a vast array of use cases. For instance, augmented reality applications track ground planes for placing virtual objects. Autonomous cars identify planar objects like traffic signs, billboards, and road markings from various distances and angles. Similarly, tasks like image stitching, robotic manipulation, and visual servoing often depend on estimating homographies between views.\\

Existing research on 3D robustness is confined to either: statistical methods, which provide weaker guarantees than formal verification~\cite{moss2023bayesian}; formal methods for 2D affine transforms~\cite{batten2024verification,singh2019abstract}, which cannot model perspective effects; or complex simulations and explicit 3D scene models~\cite{hu2024pixel}. To bridge this gap, we position homographies as a principled and analytically tractable middle ground for predominantly planar scenes (i.e. limited parallax and occlusion).

We propose the following contributions:
\begin{enumerate}
	\item \textbf{Formal verification of projective transforms:} we derive closed-form homographies induced by 6 DOF pose perturbations. We show that these parameterized homographies are compatible with recent verification work using piecewise-linear constraints and Lipschitz optimization. By extending these techniques, we derive provably tight bounds on pixel values under non-affine transforms.

	\item \textbf{Implementation and integration with existing verifiers:} we propose an efficient implementation of our method, introducing several accuracy and run time improvements to the PWL algorithm~\cite{batten2024verification}. We also adapt the \textsc{Venus} verifier to handle piecewise linear bounds. We profile and validate our implementation against PWL, demonstrating faster run times and tighter bounds.

	\item \textbf{Benchmarking and case study:} on VNN-COMP benchmarks, we reveal systematic weaknesses of some networks to 3D camera-motion perturbations. We also instantiate our method on a runway visibility classifier, illustrating how homography-based verification can reveal practical vulnerabilities in safety-critical perception, and address key regulatory certification challenges.
\end{enumerate}

\begin{figure*}[ht!]
	\centering
	\includegraphics[width=\linewidth]{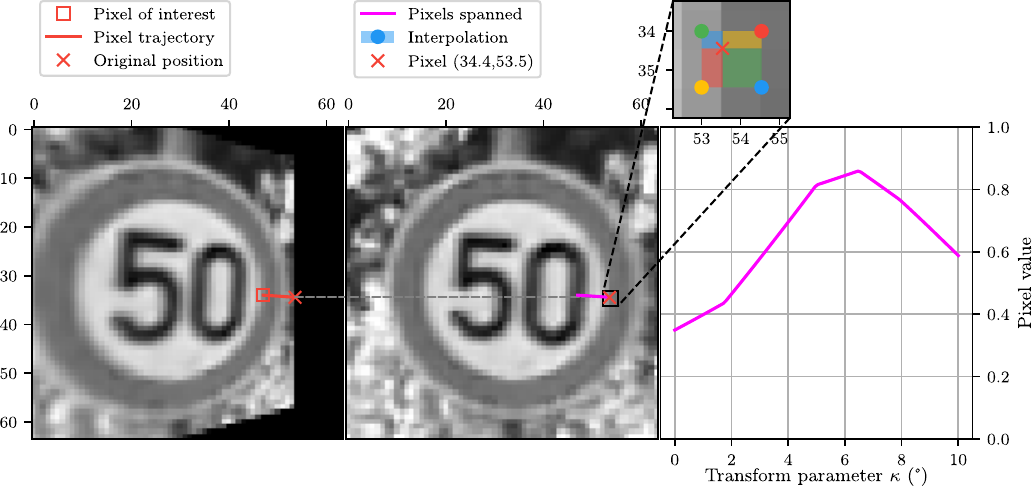}
	\begin{subfigure}[b]{0.3\textwidth}
		\centering
		\hspace*{\textwidth}
		\caption{\textbf{Transformed image}\hspace*{-1.0cm}}
		\label{fig:pixels-a}
	\end{subfigure}
	\hfill
	\begin{subfigure}[b]{0.3\textwidth}
		\centering
		\hspace*{\textwidth}
		\caption{\textbf{Original image}\hspace*{.7cm}}
		\label{fig:pixels-b}
	\end{subfigure}
	\hfill
	\begin{subfigure}[b]{0.3\textwidth}
		\centering
		\hspace*{\textwidth}
		\caption{\textbf{Pixel values}\hspace*{2.2cm}}
		\label{fig:pixels-c}
	\end{subfigure}
	\caption{\label{fig:pixels}\textbf{Pixel values under perspective transform.}
		As original image $\vect{x_0}$ \subref{fig:pixels-b} undergoes a yaw angle perturbation $\vect{\kappa} \in \notset{B}=[0\degree, 10\degree]$, we look at the values $G^{\vect{x_0}}_{34,47}(\vect{\kappa})$ of pixel $\vect{p}=(34,47)$, the red square in transformed image \subref{fig:pixels-a}. Transform $T$ is a homography and, as non-affine, is able to represent a change of perspective of the traffic sign. Applying the inverse transform, we find the corresponding position $T^{-1}(\vect{\kappa})=(34.4,53.5)$ in the original image, the red cross in \subref{fig:pixels-a} and \subref{fig:pixels-b}. We then compute the transformed pixel value with bilinear interpolation (zoomed region in \subref{fig:pixels-b}). As the yaw angle $\vect{\kappa}$ varies, pixel $\vect{p}$ follows a non-linear behavior in position (magenta path in \subref{fig:pixels-b}), and in values \subref{fig:pixels-c}: transitioning from black (low values), to white (high values), to gray. Black padding is used to highlight the transform, but other padding techniques are possible for a more realistic rendering (see supplementary material ~\cref{supp:padding}).
	}
\end{figure*}

\section{Preliminaries}
\label{sec:preliminaries}
This section provides a brief theoretical exposition to key components of our proposed method.\\

\textbf{Neural network verification.} Let $\notset{I} \triangleq [0,1]^{h \times w}$ be the set of all single-channel images of height $h$ and width $w$. Multi-channel images are handled by applying this setup to each channel independently. We consider an original image $\vect{x_0} \in \notset{I}$ which undergoes a geometric transform $T$ (\cref{fig:pixels}). The parameters $\vect{\kappa}$ of the transform are confined to intervals $\notset{B} \subseteq \RR^d$.
As $\vect{\kappa}$ spans parameter space $\notset{B}$, the set of possible images obtainable from $\vect{x_0}$ defines the \textit{attack space}:

\begin{equation}
\begin{split}
	\Omega_\notset{B}(\vect{x_0})
	\triangleq \Bigl\lbrace
		&\vect{x} \in \notset{I} ~\Big|~ x_{i,j} = G^{\vect{x_0}}_{i,j}(\vect{\kappa}); \\
		&~\forall \vect{\kappa} \in \notset{B}, \quad \forall (i,j) \in \llbracket 0,h \rrbracket \times \llbracket 0,w \rrbracket
	\Bigr\rbrace
\end{split}
\end{equation}
where $G^{\vect{x_0}}_{i,j}(\vect{\kappa})$ is the value of pixel $(i,j)$ in the image transformed from $\vect{x_0}$, by parameter $\vect{\kappa}$.
Neural network verification then over-approximates $\Omega_\notset{B}(\vect{x_0})$ using linear bounds for each pixel. To simplify notation in subsequent equations, we assume a fixed pixel $(i,j)$ in the transformed image $\vect{x}$ from $\vect{x_0}$, and drop the indexing.

To obtain a pixel value in the transformed image, the corresponding coordinates in the original image are first computed through the preimage as $(u_0,v_0) = T^{-1}(\vect{\kappa})$. Since $(u_0, v_0)$ may not correspond to exact pixel coordinates, bilinear interpolation is often applied to neighboring pixels for a better-looking result \citep[Eq.2]{batten2024verification}. The pixel value in the transformed image is thus obtained as $G(\vect{\kappa}) = I\left(u_0, v_0\right)$:

\begin{equation}
    G^{\vect{x_0}}_{i,j}(\vect{\kappa}) \triangleq G(\vect{\kappa}) = I \circ T^{-1}(\vect{\kappa})
    \label{eq:pixel-transform}
\end{equation}

Following~\cite{batten2024verification}, $G(\vect{\kappa})$ is then approximated by $q$ piecewise-linear bounds $\vect{w}, \vect{b} \in \RR^d$ as:
\begin{equation}
\begin{cases}
	\LB(\vect{\kappa}) =
	\max_{j \in \llbracket 1,q \rrbracket} \set{\tr{\vect{\underline{w}}}_j \vect{\kappa} + \vect{\underline{b}}_j}
	\leq G(\vect{\kappa}) \\
	\UB(\vect{\kappa}) =
	\min_{j \in \llbracket 1,q \rrbracket} \set{\tr{\vect{\overline{w}}}_j \vect{\kappa} + \vect{\overline{b}}_j}
	\geq G(\vect{\kappa})
\end{cases}
\label{eq:pwl_bounds}
\end{equation}
These linear bounds can then be propagated through neural network verifiers to assure that the network output is unchanged under the perturbation (e.g. ~\cite[eq.1]{batten2024verification}).\\

\textbf{Homographies.}
A homography is a projective transformation that maps lines to lines~\cite[Sec.2.3]{hartley2003multiple}. In the context of computer vision, homographies arise in various scenarios such as central projection from 3D world to a 2D camera plane (i.e. pinhole camera model), or when taking several images of the same planar surface. Using homogeneous coordinates, the homography between image points $\vect{p_0} = \tr{(u_0,v_0,1)}$ and $\vect{p}=\tr{(u,v,1)}$ can be represented as a $3 \times 3$ matrix so that $\vect{p} = \matr{H} \vect{p_0}$. In Cartesian coordinates:
\begin{equation}
\begin{cases}
	u &= \frac{H_{1,1}u_0 + H_{1,2}v_0 + H_{1,3}}{H_{3,1}u_0 + H_{3,2}v_0 + H_{3,3}} \\[1ex]
	v &= \frac{H_{2,1}u_0 + H_{2,2}v_0 + H_{2,3}}{H_{3,1}u_0 + H_{3,2}v_0 + H_{3,3}}
\end{cases}
\label{eq:homography-def}
\end{equation}
Affine transforms are a special case where $H_{3,1} = H_{3,2} = 0$ and $H_{3,3} = 1$. For general homographies, this denominator is not $1$. It represents a normalization by the depth along the optical ray, which models perspective distortions introduced by the imaging process.\\

\textbf{Lipschitz optimization.}
A real-valued function $f: \notset{B} \rightarrow \RR$ is said \textit{Lipschitz-continuous} if its gradient is bounded, intuitively it is ``limited in how fast it can change'' (see \cref{fig:lipo-bounds}): $\exists L \in \RR_+ \text{ s.t. } \forall \left(\vect{\kappa_0}, \vect{\kappa}\right) \in \notset{B}^2$,
\begin{equation}
	\abs{f(\vect{\kappa}) - f(\vect{\kappa_0})} \leq L \cdot \norm{\vect{\kappa} - \vect{\kappa_0}}_2
	\label{eq:lipschitz-continuity}
\end{equation}
This effectively bounds the values taken by the function when stepping away from a given point $\vect{\kappa_0}$:
\begin{equation}
	f(\vect{\kappa_0}) - L \cdot \norm{\vect{\kappa} - \vect{\kappa_0}}_2
	\leq f(\vect{\kappa}) \leq
	f(\vect{\kappa_0}) + L \cdot \norm{\vect{\kappa} - \vect{\kappa_0}}_2
\end{equation}
Lipschitz maximization~\cite{malherbe2017global,bachoc2021sample} leverages this property: it evaluates $f$ at $n_{\text{splits}}$ sample points in the search space, and uses the Lipschitz constant to bound potential values of the global maximum $f^\ast$ with the upper bound $\UB(x) = \min_{i \in \llbracket 1,n_\text{splits} \rrbracket} \bigl(f\left(x_i\right) + L \cdot \norm{x - x_i}_2\bigr)$ (see \cref{fig:lipo-bounds}). This process is iterated to refine the solution until a convergence criterion is met. In \citet{batten2024verification}, the procedure is adapted to find an $\varepsilon$-optimal value $\widehat{f^\ast}$ such that $\abs{f^\ast - \widehat{f^\ast}} \leq \varepsilon$.

\section{Method}
This section outlines our approach: we first model pose perturbations and derive closed-form solutions for homographies which arise from the geometry of many robotics problems. We then adapt \citet{batten2024verification} to compute piecewise-linear bounds for these non-affine transforms.

\subsection{Parameterized homographies}
We consider how ODD perturbations affect the projection of a planar scene feature (e.g. ground plane, traffic sign, road markings) onto the camera image plane (see \cref{fig:geometry}). In the scope of geometric transformations, such perturbations can be a change in camera position, orientation, uncertainty in feature location, etc. We assume the camera lens distortion known, so that we can work with undistorted images and camera rays (e.g. pinhole camera model). We also neglect parallax effects so that there is no obstruction of the scene feature by objects as the viewpoint changes.

Under these assumptions, the feature plane induces a homography between different viewpoints as the camera is moving. From \cite[Eq.13.2]{hartley2003multiple}, this homography is given by:
\begin{equation}
	\matr{H} = \matr{K} \left(\matr{R} - \frac{1}{d}\vect{t}\tr{\vect{n}} \right) \matr{K_0}^{-1}
	\label{eq:homography-vbl}
\end{equation}
with $\matr{K_0} \left[\matr{I} | \vect{0}\right]$ and $\matr{K}\left[\matr{R} | \vect{t}\right]$ the projection matrices of the first and second viewpoint, and $\vect{\pi} = \tr{\left(\tr{\vect{n}}, d\right)} \in \RR^4$ the homogeneous vector of the feature plane such that $\tr{\vect{\pi}} \vect{x} = 0$. All coordinates are expressed in the coordinate frame of the first camera. We assume the camera not aligned with the plane, ignoring the degenerate case $d = 0$.\\

We now derive the form of this homography for a plane $z = 0$ in the world frame $(W)$, thus given by $\vect{\pi}^W = \tr{(0,0,1,0)}$. We use notations from \cref{supp:notation} in the supplementary material. We assume that both viewpoints are acquired by the same camera so that $\matr{K} = \matr{K_0}$.
The plane equation becomes $0 = \tr{\left(\vect{\pi}^W\right)} \vect{x}^W = \tr{\left(\vect{\pi}^W\right)} \matr{T}^W_{C_0} \vect{x}^{C_0}$, with $\matr{T}^W_{C_0} = \left[\matr{R}^W_{C_0} | \vect{t}^W_{W \to C_0}\right]$ the passive homogeneous transform from the first camera $(C_0)$ coordinates, to the world $(W)$ coordinates.
This gives the plane equation in the first camera frame as $\tr{\left(\vect{\pi}^{C_0}\right)} \vect{x}^{C_0} = 0$ with $\vect{\pi}^{C_0} = \tr{\left(\matr{T}^W_{C_0}\right)} \vect{\pi}^W$~\cite[Eq.3.6]{hartley2003multiple}. The transform $\matr{T}^W_{C_0}$ corresponds to the first camera pose, and is parameterized by the camera Euler angles roll $\phi,\theta,\psi$ for roll, pitch, yaw, and the translation by $\vect{t}^{W}_{W \to C_0} = \tr{(x,y,z)}$. Note that the rotation angles are defined with respect to $x$-forward, $y$-right, as opposed to the typical camera frame convention with $z$-forward and $x$-right.
In the same fashion, the transform from first camera $(C_0)$ to second camera frame $(C)$ coordinates is $\matr{T}^C_{C_0} = \left[\matr{R}|\vect{t}\right] = \left[\matr{R}^C_{C_0} | \vect{t}^C_{C \to C_0}\right]$, parameterized by a delta in pose $\tr{\left(\Delta \phi, \Delta \theta, \Delta \psi, \Delta x, \Delta y, \Delta z\right)}$.
Overall, the homography $\matr{H}(\vect{\kappa})$ between two camera viewpoints is parameterized by vector $\vect{\kappa} = \tr{\left(
	\Delta \phi, \Delta \theta, \Delta \psi, \Delta x, \Delta y, \Delta z,
	\phi, \theta, \psi, x, y, z
	\right)}$, and given by \cref{eq:homography-vbl} with:
\setcounter{equation}{6}
\begin{subequations}
\begin{align}
	\matr{K} &= \matr{K_0} =
		\begin{bmatrix}
			f & 0 & x_c \\
			0 & f & y_c \\
			0 & 0 & 1
		\end{bmatrix} \\
	\matr{R} &= \matr{R}^C_{C_0}\left(\Delta \phi, \Delta \theta, \Delta \psi\right) \\
	\vect{t} &= \vect{t}^C_{C \to C_0}\left(\Delta x, \Delta y, \Delta z\right) \\
	\vect{n} &= \tr{\left(\pi_1, \pi_2, \pi_3\right)} \text{ and } d = \pi_4 \\
	\vect{\pi} &= \vect{\pi}^{C_0} = \tr{\left[\matr{T}^W_{C_0} \left(\phi, \theta, \psi, x, y, z\right)\right]} \vect{\pi}^W
\end{align}
\end{subequations}
with $f$ the camera focal length and $(x_c, y_c)$ the principal point. The full expressions of $\matr{R}^C_{C_0}$ and $\vect{t}^C_{C \to C_0}$ are given in supplementary material \cref{supp:closed-forms}.\\

We then restrict the perturbation space $\notset{B}$ to specific scenarios, which simplify the form of $\matr{H}(\vect{\kappa})$. We focus here on an example perturbation $\Delta \psi$ in yaw angle, with all other $\Delta$ parameters set to zero. The result is given in \cref{eq:yaw-deviation} and the complete derivation in supplementary material \cref{supp:yaw-deviation}. Note that since the camera center is fixed, \cref{eq:yaw-deviation} does not depend on the camera pose $\tr{\left(\phi, \theta, \psi, x, y, z\right)}$. This is a special case where $\matr{H} = \matr{K} \matr{R} \matr{K}^{-1}$ remains valid even if the projected feature is non-planar \cite[Sec.8.4.5]{hartley2003multiple}. Additional scenarios are detailed in supplementary material \cref{supp:glossary}.
\begin{equation}
\begin{split}
	&\matr{H}(\vect{\kappa}=\Delta\psi) = \\
	&\begin{bmatrix}
		\cos{\left(\Delta\psi \right)} + \frac{x_{c} \sin{\left(\Delta\psi \right)}}{f} & 0 & \frac{\left(- f^{2} - x_{c}^{2}\right) \sin{\left(\Delta\psi \right)}}{f}\\
		\frac{y_{c} \sin{\left(\Delta\psi \right)}}{f} & 1 & \frac{y_{c} \left(f \cos{\left(\Delta\psi \right)} - f - x_{c} \sin{\left(\Delta\psi \right)}\right)}{f}\\
		\frac{\sin{\left(\Delta\psi \right)}}{f} & 0 & \cos{\left(\Delta\psi \right)} - \frac{x_{c} \sin{\left(\Delta\psi \right)}}{f}
	\end{bmatrix}
\end{split}
\label{eq:yaw-deviation}
\end{equation}

\subsection{Homography verification}
With $\matr{H}(\vect{\kappa})$ known, the value of transformed pixel $\vect{p} = \tr{(u,v,1)}$ is given by the preimage and interpolation from \cref{eq:pixel-transform}. $T^{-1}$ is now the inverse of the homography \cref{eq:homography-vbl}, as opposed to an affine transform. The pixel position $T^{-1}(\vect{\kappa}) = \vect{p_0} = \tr{(u_0,v_0,1)}$ in the original image is then:
\begin{equation}
	\Bigl(u_0(\vect{\kappa}), v_0(\vect{\kappa})\Bigr)
	= \left(
		\frac{h_1(\vect{\kappa})}{h_3(\vect{\kappa})}, \frac{h_2(\vect{\kappa})}{h_3(\vect{\kappa})}
	\right)
	\label{eq:inverse_homography}
\end{equation}
with $\vect{h} = \matr{H}^{-1}(\vect{\kappa}) ~ \vect{p}$. In our example of yaw deviation \cref{eq:yaw-deviation} with $\vect{\kappa} = \Delta \psi$, this corresponds to:
\begin{equation}
\begin{cases}
		u_0(\Delta\psi) = x_c + f \cdot \frac{f \sin{\left(\Delta\psi \right)} + (u - x_c) \cos{\left(\Delta\psi \right)}}{f \cos{\left(\Delta\psi \right)} - (u - x_c) \sin{\left(\Delta\psi \right)}} \\
		v_0(\Delta\psi) = y_c + f \cdot \frac{v - y_c}{f \cos{\left(\Delta\psi \right)} - (u - x_c) \sin{\left(\Delta\psi \right)}}
\end{cases}
\label{eq:T_inv}
\end{equation}
For verification, we then over-approximate $G\left(\Delta\psi\right) = I\left[u_0(\Delta\psi), v_0(\Delta\psi)\right]$ with linear bounds that can be efficiently propagated through a neural network verifier. Given the high non-linearity of geometric transforms, especially with perspective distortions, piecewise-linear bounds enable a tighter approximation of $G(\vect{\kappa})$.\\

Following \citet{batten2024verification}, we first partition the parameter space $\notset{B} = \biguplus_{j \in \llbracket 1,q\rrbracket} \notset{B}_j$. For each subdomain, we evaluate the pixel value function at $n_s$ sample points, and derive an unsound linear bound by minimizing the error to the bound. For a lower bound, the relaxed linear program is:
\begin{equation}
	\left(\vect{\underline{w}^\ast_j},\vect{\underline{b}^\ast_j}\right) = \argmin_{\vect{\underline{w}_j},\vect{\underline{b}_j}} \left(
		\frac{1}{n_s}
		\cdot
		\sum_{\kappa_i \in \notset{B}_j} \left[G(\kappa_i) - \left(\tr{\vect{\underline{w}_j}} \kappa_i + \vect{\underline{b}_j}\right)\right]
	\right)
	\label{eq:lp}
\end{equation}
An overall unsound bound $\LB$ is then obtained by combining the piecewise segments from all subdomains (\cref{eq:pwl_bounds}). Following~\cite{batten2024verification}, we then apply Lipschitz maximization (see \cref{fig:lipo}) to cost function $\underline{J}(\vect{\kappa}) = \LB(\vect{\kappa}) - G(\vect{\kappa})$, which represents the bound violation.
By finding an analytical $\epsilon$-majorant $\widehat{\underline{J}^\ast}$, we can guarantee $\forall \vect{\kappa}, ~\underline{J}(\vect{\kappa}) \leq \widehat{\underline{J}^\ast} + \epsilon \Leftrightarrow \forall \vect{\kappa}, ~\LB(\vect{\kappa}) - \left(\widehat{\underline{J}^\ast} + \epsilon\right) \leq G(\vect{\kappa})$. The shifted bound $\LB^\ast(\vect{\kappa}) = \LB(\vect{\kappa}) - \left(\widehat{\underline{J}^\ast} + \epsilon\right)$ thus constitutes a \textit{sound} lower bound to $G$. A similar procedure is followed for the upper bound. \citet{batten2024verification} showed that for a given number of splits, the derived bounds are optimal, irrespective of the downstream robustness task.\\

\newcommand{\lipoalgorithm}{
	\begin{algorithmic}[1]
		\Require $J$, $\notset{B}$, $\epsilon > 0$
		\Ensure estimated max. $\widehat{J^\ast}$ with certificate $\epsilon$

		\State $J_{\text{max}} \gets 0$
		\State $\family{\notset{B}_i}{i \in \llbracket 1, n_{\text{splits}} \rrbracket} \gets \text{split}\left(\notset{B}\right)$

		\While{$\card{\notset{B}} > 0$}
		\State $\notset{B}_j \gets \text{pop}\left[\family{\notset{B}_i}{i}\right]$
		\State $\notset{J}_{\text{samples}} \gets \text{sample}\left(J, \notset{B}_j\right)$
		\State $J_{\text{max}} \gets \max\bigl(\set{J_{\text{max}}} \cup \notset{J}_{\text{samples}}\bigr)$
		\State $L \gets \sup_{\vect{\kappa} \in \text{Diff}\left(\notset{B}\right)} \abs{\tr{\nabla}_{\vect{\kappa}} J\left(\vect{\kappa}\right) \cdot e}$
		\State $\left[\vect{\kappa_0}, \vect{\kappa_n}\right] \gets \notset{B}_i$
		\State $J_{\text{bound}} \gets \frac{J\left(\vect{\kappa_0}\right)}{2} + \frac{J\left(\vect{\kappa_n}\right)}{2} + \frac{L}{2} \cdot \norm{\vect{\kappa_n} - \vect{\kappa_0}}$

		\If{$J_{\text{bound}} > J_{\text{max}} + \epsilon$}
		\State $\notset{B} \gets \notset{B} ~\cup~ \text{split}\left(\notset{B}_j\right)$
		\EndIf
		\EndWhile

		\State \Return $\widehat{J^\ast} = J_{\text{max}} + \epsilon$
	\end{algorithmic}
}

\begin{figure}
	\centering
	\begin{subfigure}[b]{\linewidth}
		\centering
		\par\noindent\rule{\textwidth}{1pt}
		\small
		\lipoalgorithm
		\par\noindent\rule{\textwidth}{1pt}
		\caption{Algorithm}
		\label{alg:bab-lipo}
	\end{subfigure}
	\hfill
	\begin{subfigure}[b]{\linewidth}
		\centering
		\includegraphics[width=\linewidth]{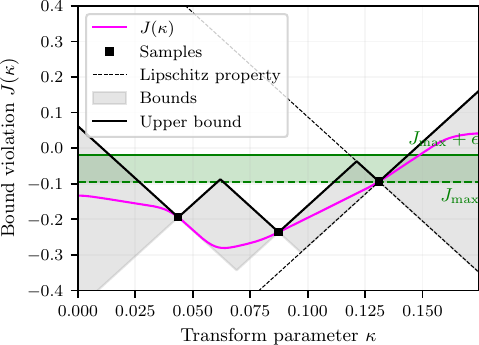}
		\caption{Visualization}
		\label{fig:lipo-bounds}
	\end{subfigure}
	\caption{\textbf{Lipschitz optimization.} Adapted from \cite{batten2024verification}, procedure \ref{alg:bab-lipo} finds an $\epsilon$-maximum of function $J$. On \cref{fig:lipo-bounds}, a few samples (black) approximate the true function (magenta). The Lipschitz property of $J$ constrains, with slope $\pm L$, possible values for the function (gray bands). In this example, since the maximum of the Lipschitz bound (black) is above $J_{\max} + \epsilon$, the interval is split a few more times to better estimate $J_{\max}$, and capture a $\epsilon$-optimum.}
	\label{fig:lipo}
\end{figure}

The missing argument is then to find a valid upper bound to the Lipschitz constant. In~\cite{batten2024verification}, it is estimated with the directional gradient as $L_m = \sup_{\vect{\kappa} \in \diff\left(\notset{B}\right)} \abs{\tr{\nabla}_{\vect{\kappa}} J\left(\vect{\kappa}\right) \cdot e_m}$ for coordinate $m$. We extend this formula in the case of homographies where the denominator $h_3(\vect{\kappa})$ may render $G$ non-continuous (\cref{eq:inverse_homography}) and break its Lipschitz properties. With additional scenarios developed in supplementary material \cref{supp:glossary}, our yaw perturbation gradient from \cref{eq:yaw-deviation} is:
\begin{equation}
	\tr{\nabla_{\Delta\psi}} T^{-1}\left(\Delta\psi\right) = \begin{pmatrix}
		\frac{f \left[f^{2} + (u - x_c)^2\right]}{\left[f \cos{\left(\Delta\psi \right)} - (u - x_c) \sin{\left(\Delta\psi \right)}\right]^{2}}\\[1.5ex]
		\frac{f (v - y_c) \left[f \sin{\left(\Delta\psi \right)} + (u - x_c) \cos{\left(\Delta\psi \right)}\right]}{\left[f \cos{\left(\Delta\psi \right)} - (u - x_c) \sin{\left(\Delta\psi \right)}\right]^{2}}
	\end{pmatrix}
	\label{eq:grad_T}
\end{equation}
From \cref{eq:T_inv} we first notice that $T^{-1}$ is not continuous at the critical angle ${\Delta\psi}_c = \arctan\left(\frac{f}{u - x_c}\right)$ if $u \neq x_c$, and $\frac{\pi}{2}$ otherwise. Thus $\diff(\notset{B}) = \notset{B} \setminus \set{{\Delta\psi}_c \pmod{\pi}}$. For the scope of small yaw perturbations around $\Delta\psi = 0\degree$, the pixel-dependent critical angle does not belong to the parameter space $\notset{B}$. We also note that for diverging pixel coordinates at ${\Delta\psi}_c$, the pixel function $G$ would remain bounded as the interpolation uses padding values. In practice, we found $\abs{{\Delta\psi}_c} \in \left[72\degree,90\degree\right] \not\subset \notset{B}$. With $\diff(\notset{B})$ now defined, we derive candidate locations for maximum gradient values (proof in supplementary material \cref{supp:gradient-derivation}). From \cref{eq:grad_T}, and for $\notset{B} = \left[\Delta\psi_{\min}, \Delta\psi_{\max}\right]$:
\begin{align}
	\begin{split}
		&\argmax_{\Delta\psi \in \diff(\notset{B})}
		\Bigl|\nabla_{\Delta\psi} u_0\left(\Delta\psi\right)\Bigr| \in \\
		&\set{
			\Delta\psi_{\min},~
			\Delta\psi_{\max},~
			\arctan\left(\frac{x_c - u}{f}\right) \bmod{\pi}}\\
	\end{split}
	\\
	&\argmax_{\Delta\psi \in \diff(\notset{B})}\Bigl|\nabla_{\Delta\psi} v_0 \left(\Delta\psi\right)\Bigr| \in
	\set{\Delta\psi_{\min}, \Delta\psi_{\max}}
\end{align}

For the gradient of the discontinuous interpolation function, we use the maximum gradient from the different interpolation regions~\cite{balunovic2019certifying,batten2024verification}. Finally, remembering our notation simplification $G = G^{\vect{x_0}}_{i,j}$, the process is repeated for each pixel $(i,j)$ to obtain bounds for all pixels of the transformed image. These bounds are then passed to a verifier to guarantee the network output is unchanged under the perturbation.

\section{Evaluation}
This section introduces our experimental setup and results. Experiments are performed on a laptop using an Intel i7-13800H processor (20 threads, 5GHz), 32GB RAM, and a NVIDIA RTX 2000 GPU (driver 550.120, CUDA 12.4).

\textbf{Datasets.} We evaluate our method on the VNN competition benchmarks~\cite{brix2023vnncomp}. We choose datasets of increasing complexity (model and image size) that are relevant to robotics applications: digit classification (MNIST), object recognition (CIFAR-10), and autonomous driving (GTSRB). We use 100 images from each test set, and estimate the focal length to yield reasonable angular perturbations that are similar between datasets (see supplementary material~\cref{supp:glossary}). Additionally, we use a runway visibility classifier developed on the LARD dataset~\cite{ducoffe2023lard}.

\textbf{Verification.} Our bounds are solver-agnostic, decoupling transformation modeling (front end) from the network verification engine (back end). We choose \textsc{Venus} as the only baseline supporting a MILP formulation for \textit{complete} robustness of piecewise-linear bounds. We enhanced its bound-propagation framework to derive pre-activation bounds from linear input constraints rather than fixed intervals. This enables {\sc Venus} to construct tighter ReLU relaxations and better constrain the underlying MILP problem.

\textbf{Metrics.} First, we evaluate our method independently of any downstream verification task: polytope tightness, number of branch-and-bound steps (BaB), and run time per image. Then, we use our bounds to evaluate model robustness: we measure average run times, and number of robust cases (i.e. a correct model output is unchanged by the transform).

\begin{figure}
	\centering
	\includegraphics[width=\linewidth]{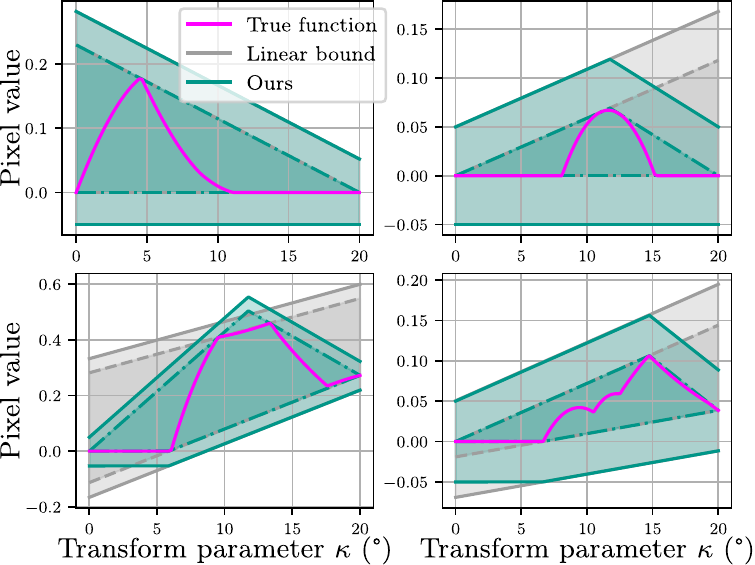}
	\caption{\label{fig:bounds}
		\textbf{Example bounds.} Pixel values (magenta) are over-approximated. Unsound bounds (dashed) are shifted to sound bounds (solid) by Lipschitz optimization. Depending on complexity, curves have zero (linear), one (PWL~\cite{batten2024verification}), or two splits (ours).
	}
\end{figure}

\textbf{Validation.} Analytical Lipschitz constants were validated with finite differences by oversampling the pixel curve. Unit tests validated \textsc{Venus}~\cite{botoeva2020efficient} modifications for piecewise-linear bounds. For re-implementing PWL, we profiled the code and enforced bit-identical results at multiple stages of the algorithm, before adding our modifications.

\subsection{Comparison to \citet{batten2024verification}}
We re-implement PWL~\cite{batten2024verification} to handle perspective transforms by adding closed-form homographies, associated gradients, and Lipschitz constants. Accuracy and exactness are improved through algorithmic changes: 1) half-pixel corrections for exact gradients; 2) inclusion of omitted $\vect{\kappa}$ values in $\diff(\notset{B})$ to yield extra Lipschitz constant candidates; 3) dichotomy on $\notset{B}$ from the first iteration, reducing branch-and-bound steps; 4) tighter bounds by using $q=2$ for both upper and lower bounds; and 5) parallelization at the pixel level. Since PWL supports only affine transforms, we compare implementations under 2D rotations (\cref{tab:comparison_batten}).

\begin{table}[ht!]
	\centering
	\small
	\setlength{\aboverulesep}{0pt}
	\setlength{\belowrulesep}{0.5pt}
	\setlength{\tabcolsep}{4pt}
	\renewcommand{\arraystretch}{0.9} 
	\renewcommand{\theadfont}{\small\bfseries}

	\begin{tabular}{l cc cc}
		\toprule
		& \multicolumn{2}{c}{\textbf{MNIST}} & \multicolumn{2}{c}{\textbf{CIFAR-10}} \\
		\cmidrule(lr){2-3} \cmidrule(lr){4-5}
		\textbf{5 degrees} & \thead{PWL} & \thead{Ours} & \thead{PWL} & \thead{Ours} \\
		\midrule
			\quad \textbf{Polytope area} ($\small\times 10^{-3}$) $\downarrow$        & 9.48 & \textbf{9.42} & 10.94 & \textbf{10.83} \\
			\quad \textbf{BaB steps} $\downarrow$            & 124.0 & \textbf{17.5} & 124.0 & \textbf{35.5} \\
			\quad \textbf{Time (s)} $\downarrow$             & 87.6 & \textbf{7.0} & 910.3 & \textbf{98.7} \\
			\quad \textbf{Time @ fixed BaB (s)} $\downarrow$ & 87.6 & \textbf{12.9} & 910.3 & \textbf{148.1} \\

		\addlinespace

		\textbf{20 degrees} \\
		\midrule
			\quad \textbf{Polytope area} ($\small\times 10^{-2}$) $\downarrow$ & 9.33 & \textbf{8.81} & 15.1 & \textbf{14.0} \\
			\quad \textbf{BaB steps} $\downarrow$            & 137.0 & \textbf{72.5} & 124.2 & \textbf{90.5} \\
			\quad \textbf{Time (s)} $\downarrow$             & 90.3 & \textbf{36.2} & 891.0 & \textbf{530.9} \\
			\quad \textbf{Time @ fixed BaB (s)} $\downarrow$ & 90.3 & \textbf{24.0} & 891.0 & \textbf{415.2} \\
			\bottomrule
	\end{tabular}
	\caption{\label{tab:comparison_batten}
		\textbf{Average performance comparison with PWL~\citep{batten2024verification}.}
	}
\end{table}

High run times are intrinsic to \textit{complete} branch-and-bound verifiers; no cheaper alternative provides complete certificates to piecewise-linear bounds. Our implementation reduces BaB iterations by more than 71\% on small transforms, for an 89\% total speedup. On larger transforms, this fades to 40\% as the overhead of splitting non-converging candidates becomes dominant. At a fixed BaB budget, our per-pixel parallelization is still 53-85\% faster. The choice of $q=2$ yields marginally tighter bounds for small transforms, and up to 7\% on more complex pixel curves (\cref{fig:bounds}).

\subsection{VNN competition benchmarks}
We perform the first robustness evaluation of the VNN benchmark networks~\cite{brix2023vnncomp} against perspective transforms. We generate and propagate bounds through \textsc{Venus} to verify network robustness against the benchmark specifications (\cref{tab:overall}). We use black padding, a Lipschitz-error of $0.01$, and $5,000$ maximum BaB iterations.

\begin{table}[ht!]
	\centering
	\small
	\settowidth{\rotheadsize}{\small\bfseries MetaRoom}
	\setlength{\aboverulesep}{0pt}
	\setlength{\belowrulesep}{0pt}
	\setlength{\tabcolsep}{5pt}
	\renewcommand{\arraystretch}{0.9} 
	\renewcommand{\theadfont}{\small\bfseries}

	\begin{tabular}{ll ccc}
		\toprule
			\thead{Perturbation}
			& \thead{$\bm{\notset{B}}$}
			& \thead{MNIST}
			& \thead{CIFAR-10}
			& \thead{GTSRB} \\
		\midrule
			$\bm{\Delta \phi}$ (roll)    & $\bm{[0, 5]}^\circ$   & 61 & 69 & 0 \\
			$\bm{\Delta \theta}$ (pitch) & $\bm{[0, 5]}^\circ$   & 5  & 10 & 0 \\
			$\bm{\Delta \psi}$ (yaw)     & $\bm{[0, 5]}^\circ$   & 25 & 7  & 0 \\
			$\bm{\Delta x}$              & $\bm{[0,1]~m}$ & 50 & 21 & 0 \\
			$\bm{\Delta y}$              & $\bm{[0,1]~m}$ & 73 & 83 & 0 \\
			$\bm{\Delta z}$              & $\bm{[0,1]~m}$ & 53 & 24 & 0 \\
		\midrule
			\multicolumn{2}{l}{\textbf{Generation time (s)} $\downarrow$} & 42.0 & 1858.6 & 292.6 \\
			\multicolumn{2}{l}{\textbf{Verification time (s)}
        $\downarrow$} & 8.23 & 41.33 & 5323.8 \\
			\multicolumn{2}{l}{\textbf{Total timeouts}
$\downarrow$} & 0 & 2 & 178 \\
		\bottomrule
	\end{tabular}
	\caption{\label{tab:overall}\textbf{Ratio (\%) of robust cases to non-affine perturbations.}}
\end{table}

\cref{tab:overall} shows that even PGD-trained networks can be highly vulnerable to 3D geometric perturbations. MNIST and CIFAR models are less robust to pitch, yaw, $\Delta x$ and $\Delta z$ transforms than to the $\ell_p$-norm perturbations reported in~\cite{bak2021second}. Roll and $\Delta y$ perturbations appear more robust since, under our assumptions, they correspond exactly to affine transforms (rotation and shear, \cref{supp:glossary}) better approximated by $\ell_p$-norm attacks and commonly included in data augmentation. More surprisingly, the autonomous-driving models exhibit no certified robustness to 3D perturbations on GTSRB, reinforcing prior evidence that these networks lack robustness even to simpler $\ell_p$-norm attacks~\cite{brix2023vnncomp,kaulen2025vnncomp}.

\subsection{Case study: runway visibility}
Following current aviation safety recommendations~\cite{faa2024roadmap,easa2024guidance}, we illustrate our method on a low-criticality application: a binary classifier that predicts runway visibility, informing the pilot’s decision to continue the approach or initiate a go-around.
The network was trained on LARD~\cite{ducoffe2023lard} without robustness objectives.
Applying our homography-based verification, we found limited robustness: only 16\% of test images were certified under 10 cm translations and 1\% under 1° rotations. This underscores the difficulty of achieving robustness to camera motion even in simple classifiers, and motivates future work on robust training in this setting.

\subsection{Sensitivity and limitations}
\textbf{Amplitude.} Large transforms exacerbate the non-linearities of the pixel curve and complicate its approximation by piecewise-linear bounds. \cref{tab:amplitude} shows how certified robustness decreases with amplitude. This is a well-known challenge in formal verification, typically addressed through \textit{input splitting} \cite{xu2021fast,wu2024marabou} by partitioning the domain into smaller intervals: a near-linear regime precisely consistent with homography modeling and our planar assumptions.

\begin{table}[ht!]
	\centering
	\small
	\setlength{\aboverulesep}{0pt}
	\setlength{\belowrulesep}{0pt}
	\setlength{\tabcolsep}{4pt}
	\renewcommand{\arraystretch}{0.9} 
	\renewcommand{\theadfont}{\small\bfseries}

	\begin{tabular}{lcccccccc}
		\toprule
		\textbf{Yaw angle} & $\bm{1\degree}$ & $\bm{2\degree}$ & $\bm{3\degree}$ & $\bm{4\degree}$ & $\bm{5\degree}$ & $\bm{10\degree}$ & $\bm{15\degree}$ & $\bm{20\degree}$ \\
		\midrule
		\textbf{Verified (\%)} & 76 & 68 & 57 & 41 & 25 & 0 & 0 & 0 \\
		\bottomrule
	\end{tabular}
	\caption{\label{tab:amplitude}\textbf{Robustness to yaw amplitude on MNIST.}}
\end{table}

\textbf{Padding.} Bilinear interpolation uses a padding value to fill in missing pixels transformed from outside the original image bounds (\cref{fig:pixels}). \cref{tab:padding} shows how padding that duplicates image content offers better robustness than padding with stark borders, which induce distribution shift. While black padding remains standard in benchmarks and aligns with CIFAR/MNIST training augmentations, it risks exposing GTSRB models to border artifacts (e.g., vignetting) rather than geometric shifts. However, substituting realistic reflection padding yields unchanged experiment results, reiterating the known brittleness of these networks \cite{brix2023vnncomp,kaulen2025vnncomp}.

\begin{table}[ht!]
	\centering
	\small
	\setlength{\aboverulesep}{0pt}
	\setlength{\belowrulesep}{0pt}
		\setlength{\tabcolsep}{4pt}
	\renewcommand{\arraystretch}{0.9} 
	\renewcommand{\theadfont}{\small\bfseries}

	\begin{tabular}{cccccc}
		\toprule
		$\bm{\notset{B}}$ & \textbf{Black} & \textbf{Gray} & \textbf{Replicate} & \textbf{Reflect} & \textbf{Wrap} \\
		\midrule
		$\bm{[0,2.5]\degree}$ & 34 & 33 & 42 & 42 & 35 \\
		$\bm{[0,5]\degree}$ & 7 & 7 & 8 & 9 & 7 \\
		\bottomrule
	\end{tabular}
	\caption{\label{tab:padding}\textbf{Robustness to yaw padding on CIFAR-10.}}
\end{table}

\section{Related Work}
We focus on the verification of pre-trained models, without considering \textit{robust-training} techniques \cite{yang2023provable}.

\smallskip

\textbf{Robustness of neural networks.}
Neural networks, while capable of approximating complex functions, exhibit inherent instability stemming from repeated compositions of linear and non-linear operations~\citep{szegedy2014intriguing, goodfellow2014explaining}. This vulnerability to input perturbations raises significant concerns about their reliability in safety-critical applications in the real-world, where inputs are subject to noise, distortions, or adversarial manipulations. Verification thus aims to provide robustness guarantees against input perturbations~\citep{liu2021algorithms,huchette2026when,li2023sok}. Without loss of generality, most methods focus on the feed-forward fully-connected architecture~\citep{huchette2026when}.

\smallskip

\textbf{\boldmath$\ell_p$-norm perturbations.}
In this context, robustness to $\ell_p$-norm perturbations is often modeled as an optimization problem~\citep{li2023sok}.
\textit{Incomplete} verifiers scale to large vision models, but may not always provide conclusive answers.
In contrast, \textit{complete} verifiers such as Reluplex~\citep{katz2017reluplex}, Marabou~\citep{wu2024marabou}, or \textsc{Venus}~\citep{botoeva2020efficient} focus on exact formulations through satisfiability modulo theory (SMT) or mixed-integer linear programming (MILP). They provide definitive satisfiability answers but are limited to small networks.

Interval bound propagation (IBP), used in DeepPoly~\cite{singh2019abstract} and Lipschitz-based methods~\cite{batten2024verification}, offers faster verification by relaxing the ReLU polytope, but often over-approximates the underlying optimisation constraints. Probabilistic methods~\cite{moss2023bayesian} scale well and can handle arbitrary \textit{black-box} architectures, but provide weaker, probabilistic guarantees than formal methods. The Branch-and-bound (BaB) paradigm combines complete and incomplete verification, thereby scaling to large models while still providing definitive robustness answers. BaB is the current state of the art, with $\alpha$-$\beta$-CROWN~\cite{zhang2018efficient,xu2020automatic,xu2021fast,wang2021beta,shi2025neural,zhang2022general} dominating the past few VNN-COMP leaderboards~\cite{brix2023vnncomp}.

Despite significant progress in $\ell_p$-norm verification, recent work has highlighted its limitations in capturing the complexity of real-world perturbations~\cite{mangal2023certifying}. As illustrated in \cref{fig:manifold}, the space of possible perturbations extends far beyond the confines of $\ell_p$-balls~\citep{engstrom2019exploring,dai2024position}: pixel flow~\citep{xiao2018spatially}, data augmentation~\citep{gao2020fuzz}, multi-norm~\citep{jiang2024towards}, universal adversarial perturbations (UAPs)~\citep{moosavi2017universal}, distribution~\citep{wu2023toward}, Wasserstein $W_p$-balls~\citep{sinha2017certifying}, composition~\citep{hsiung2023towards}, etc. In particular, $\ell_p$-bounded attacks cannot directly model geometric perturbations such as changes of the camera viewpoint.

\smallskip

\textbf{Geometric perturbations.} Through non-linear mappings of pixel coordinates, geometric transformations exacerbate the inherent non-linearity of neural networks, and their vulnerability to input variations~\citep{engstrom2019exploring,fawzi2017robustness,kanbak2018geometric,li2023sok}. While the set of rotations of an image does not form an $\ell_p$-ball, a common approach consists in bounding the rotation parameters to retrieve a problem formulation that is compatible with $\ell_p$-norm verifiers~\citep{singh2019abstract,wang2022art,yang2023provable}.
\citet{balunovic2019certifying} linked geometric parameter to pixel value bounds but for 2D transforms only. They also contributed a method for verifying bilinear interpolation, commonly used with geometric transforms. \citet{batten2024verification} achieved tighter bounds by employing piecewise-linear approximation and Lipschitz optimization, again for 2D transforms. Their approach generalizes to matrix-defined affine transforms~\citep{dumont2018robustness,kouvaros2018formal,balunovic2019certifying,hsiung2023towards}.

Beyond affine transforms, \citet{hu2022robustness,hu2023robustness,hu2024pixel} verify models against camera motion perturbations with pixel-wise smoothing, but require a point cloud of the scene. The Lipschitz procedures from DeepGO \citep{ruan2018reachability} and GeoRobust \citep{wang2023towards} are used to analyze the robustness \textit{of the network}, rather than for linearly approximating the input bounds as we propose.
ManiFool~\citep{kanbak2018geometric} models projective transforms by traversing the perturbation manifold to find counter-examples. However, the resulting metric is difficult to compare with standard robustness specifications (e.g., class score margin). \citet{hanspal2023efficient} abstract complex 3D transforms through an encoding head, and verify specifications in the resulting latent space. This approach is very efficient but modifies the network, and does not leverage potential closed-form solutions.

\smallskip

\textbf{Homographies in safety-critical applications.}
Homographies are central to many robotics and vision systems that exploit planar geometry. In autonomous driving, inverse perspective mapping (IPM) uses a ground-plane homography to generate bird’s-eye-view representations for lane detection and scene understanding~\citep{bruls2019right,ma2024vision}. In robotic surgery, homographies support calibration to align robot-mounted cameras with the workspace~\citep{tsai1989new}, and augmented reality to overlay anatomical models from endoscopic or X-ray views~\citep{chen2017real}. Finally, for autonomous drones, vision systems often detect a known planar pattern (e.g., an AprilTag marker or helipad) and use homographies to estimate the vehicle’s relative pose for landing~\citep{sefidgar2022landing, chavez2017homography}.

\section{Conclusion}
Moving beyond the limitations of $\ell_p$-norm and affine robustness, we propose the first general framework to formally verify 3D projective transforms, with applications across core computer-vision and robotics settings (e.g. augmented reality, autonomous vehicles, and manipulation). Our work derives closed-form homographies from 6-DOF camera pose and extends the PWL algorithm to compute provably tight pixel bounds, achieving up to 89\% faster run times and 7\% tighter bounds. Our experiments reveal previously unreported vulnerabilities of benchmark networks to 3D camera motions. Finally, a case study on a runway classifier illustrates how homography-based verification can be applied to perception modules in aerospace, marking progress towards the certification of AI-powered systems in safety-critical applications.

\newpage
\section*{Acknowledgments}
The authors would like to thank Ben Batten for helpful discussions on the PWL algorithm and the \textsc{Venus} verifier.

{
    \small
    \bibliographystyle{ieeenat_fullname}
    \bibliography{references}
}

\onecolumn
\setcounter{page}{1}

\renewcommand{\maketitlesupplementary}{
   \newpage
   \begin{center}
		\Large
		\textbf{\thetitle}\\
		\vspace{0.5em}Supplementary Material \\
		\vspace{1.0em}
   \end{center}
}

\maketitlesupplementary

\section{Notation}
\label{supp:notation}

\begin{tabular}{cl}
	$\triangleq$ & Equality by definition \\
	$\llbracket 1,n \rrbracket$ & Set of natural integers between $1$ and $n$, both included \\
	$\notset{A,B,C,\dotsc}$ & Sets (cursive script) \\
	$\diff\left(\notset{B}\right)$ & Subset of $\notset{B}$ where the bound error $\underline{f}$ is differentiable \\
	$f^\ast$ & Optimum value (maximum or minimum) of function $f$ \\
	$\widehat{f^\ast}$ & Estimate of $f^\ast$ \\
	$G^{\vect{x_0}}_{i,j}(\vect{\kappa})$ & Value of pixel $(i,j)$ in the transformed image, when image $\vect{x_0}$ is transformed by $\vect{\kappa}$\\
	$\vect{x}$ & Vectors, with $1$-indexed elements $x_i$ (bold lowercase) \\
	$\vect{x}^A$ & Vector $\vect{x}$, expressed in coordinate frame $A$ \\
	$\matr{A}$ & Matrix, with $1$-indexed elements $A_{i,j}$ (bold uppercase) \\
	$(A)$ & The $(A)$ coordinate frame \\
	$\matr{R}^B_A$ & $3 \times 3$ passive rotation from $A$-frame to $B$-frame coordinates, so that $\vect{x}^B = \matr{R}^B_A \vect{x}^A$ \\
	$\vect{t}^B_{B \to A}$ & $3 \times 1$ translation vector from $B$-frame to $A$-frame, expressed in frame $B$ \\
	$\matr{T}^B_A$ & $4 \times 4$ passive transform from $(A)$ to $(B)$, so $\vect{x}^B = \matr{T}^B_A \vect{x}^A$. Note $\matr{T}^B_A = \left[\matr{R}^B_A | \vect{t}^B_{B \to A}\right]$ \\
	$[\matr{R} | \vect{t}]$ & Simplified form of the $4 \times 4$ homogeneous coordinate transform $\left[
	\begin{array}{@{}c|c@{}}
		\matr{R}     & \vect{t} \\ \hline
		\begin{array}{@{}ccc@{}}
			0 & 0 & 0
		\end{array}  & 1
	\end{array}
	\right]$
\end{tabular}

\section{Datasets}
\begin{figure}[ht!]
	\centering
	\includegraphics[width=0.5\linewidth]{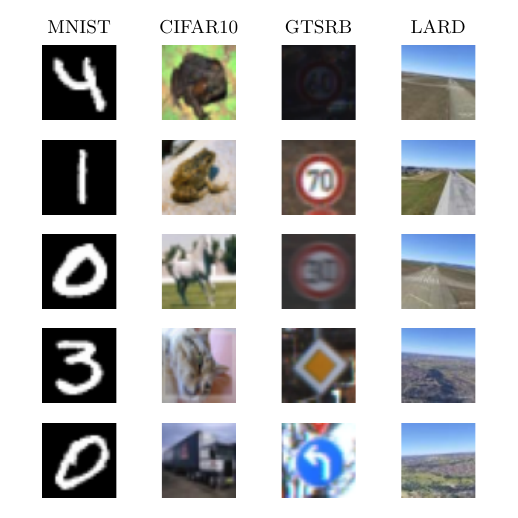}
	\caption{\label{fig:datasets} Sample images from each dataset.}
\end{figure}

\section{Padding}
\label{supp:padding}
\begin{figure}[ht!]
	\centering
	\includegraphics[width=\linewidth]{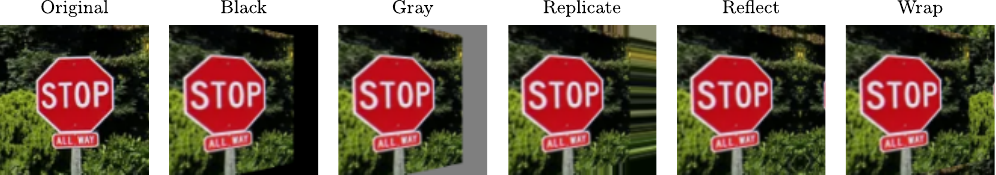}
	\caption{\label{fig:padding} Overview of different padding techniques.}
\end{figure}

Geometric transformations frequently map target coordinates beyond the boundaries of the source image. In this case, bilinear interpolation relies on padding mechanisms to impute missing pixel values. Depending on the application, different padding techniques exist, with varying levels of complexity and realism. For the traffic sign of \cref{fig:padding}, ``reflect'' padding offers a perceptually plausible extension of the background texture. Note how padding introduces non-existent signal into the image, which is distinct from having a true change of camera perspective with for example, occlusion effects.

\section{Transforms glossary}
\label{supp:glossary}
We detail a few scenarios that simplify the general homography form given in \cref{eq:homography-vbl}. Under our assumptions, we show that some of these scenarios simplify into exact affine transforms. We also notice that rotation perturbations are independent of the pose of the initial viewpoint. For brevity of the translation cases, we simplify further by assuming a camera initially aligned with the world axes ($\phi,\theta,\psi = 0$). We keep $x,y,z \neq 0$, and notice that the homography does not depend on the initial camera longitudinal or lateral position. It depends however on the delta in position $\Delta x, \Delta y, \Delta z$. To explicitly highlight the effects of the transform on the image, our figures utilize black padding. However, we refer to \cref{supp:padding} for padding techniques that prioritize perceptual realism. The more complete forms are available with our public code.

\subsection{Roll perturbation \texorpdfstring{$\bm{\Delta\phi}$}{dphi}}
\begin{figure}[ht!]
	\includegraphics[width=\linewidth]{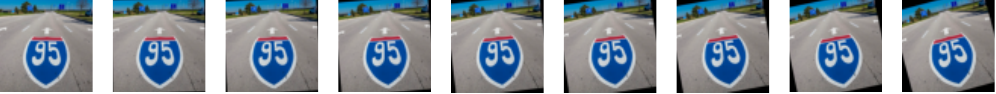}
	\caption{Samples for a roll perturbation, with $\vect{\kappa}=\Delta\phi \in \notset{B} = [0,15]\degree$}
\end{figure}

We notice that under our assumptions, the roll perturbation is an affine transform and corresponds exactly to a 2D rotation of the image. It is also independent of the camera focal length $f$.
\begin{align}
	\matr{H}^{-1}(\Delta\phi)  &= \begin{bmatrix}
		\cos{\left(\Delta\phi \right)} & - \sin{\left(\Delta\phi \right)} & - x_{c} \cos{\left(\Delta\phi \right)} + x_{c} + y_{c} \sin{\left(\Delta\phi \right)}\\
		\sin{\left(\Delta\phi \right)} & \cos{\left(\Delta\phi \right)} & - x_{c} \sin{\left(\Delta\phi \right)} - y_{c} \cos{\left(\Delta\phi \right)} + y_{c}\\
		0 & 0 & 1
	\end{bmatrix} \\
	\Bigl(u_0(\Delta\phi), v_0(\Delta\phi)\Bigr) &= \begin{pmatrix}
			x_c + (u - x_c) \cos(\Delta\phi) - (v - y_c) \sin(\Delta\phi),~\\
			y_c + (u - x_c) \sin(\Delta\phi) + (v - y_c) \cos(\Delta\phi)
		\end{pmatrix}\label{eq:roll_u0_v0}\\
	\tr{\nabla_{\Delta\phi}} T^{-1}(\Delta\phi) &= \begin{pmatrix}
		- (u - x_c) \sin(\Delta\phi) - (v - y_c) \cos(\Delta\phi),~\\
		(u - x_c) \cos(\Delta\phi) - (v - y_c) \sin(\Delta\phi)
	\end{pmatrix}\\
	\diff(\notset{B}) &= \notset{B} \\
	\argmax_{\Delta\phi \in \diff(\notset{B})}
	\Bigl|\nabla_{\Delta\phi} J\left(\Delta\phi\right)\Bigr| &\in
	\set{{\Delta\phi}_{\min}, {\Delta\phi}_{\max}, \arctan\left(-\frac{v - y_c}{u - x_c}\right), \arctan\left(\frac{u - x_c}{v - y_c}\right)} \pmod{\pi}
\end{align}
The limit cases of $\arctan$ apply when $u = x_c$ and $v = y_c$.

\subsection{Pitch perturbation \texorpdfstring{$\bm{\Delta\theta}$}{dtheta}}
\begin{figure}[H]
	\includegraphics[width=\linewidth]{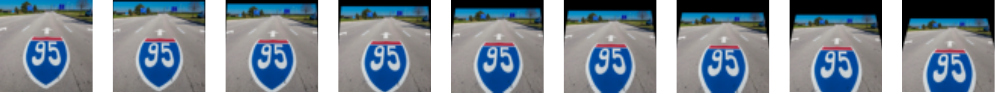}
	\caption{Samples for a pitch perturbation, with $\vect{\kappa}=\Delta\theta \in \notset{B} = [0,15]\degree$}
\end{figure}

\begin{align}
	&\matr{H}^{-1}(\Delta\theta)=
		\begin{bmatrix}
			1 & \frac{x_{c} \sin{\left(\Delta\theta \right)}}{f} & \frac{x_{c} \left(f \cos{\left(\Delta\theta \right)} - f - y_{c} \sin{\left(\Delta\theta \right)}\right)}{f} \\
			0 & \cos(\Delta\theta) + \frac{y_c \sin(\Delta\theta)}{f} & -\frac{\left(f^{2} + y_{c}^{2}\right) \sin{\left(\Delta\theta \right)}}{f} \\
			0 & \frac{\sin{\left(\Delta\theta \right)}}{f} & \cos{\left(\Delta\theta \right)} - \frac{y_{c} \sin{\left(\Delta\theta \right)}}{f}
		\end{bmatrix} \\
	&\Bigl(u_0(\Delta\theta), v_0(\Delta\theta)\Bigr) = \begin{pmatrix}
		x_c + f \cdot \frac{u - x_c}{f\cos(\Delta\theta) + (v - y_c) \sin(\Delta\theta)},~
		y_c - f \cdot \frac{f\sin(\Delta\theta) - (v - y_c)\cos(\Delta\theta)}{f\cos(\Delta\theta) + (v - y_c) \sin(\Delta\theta)}
	\end{pmatrix}\\
	&\tr{\nabla_{\Delta\theta}} T^{-1}(\Delta\theta) = \begin{pmatrix}
		- \frac{f (u - x_c)\left[- f\sin(\Delta\theta) + (v - y_c)\cos(\Delta\theta)\right]}{\left[f\cos(\Delta\theta) + (v - y_c) \sin(\Delta\theta)\right]^2},~
		\frac{-f \left[f^2 + (v - y_c)^2\right]}{\left[f\cos(\Delta\theta) + (v - y_c) \sin(\Delta\theta)\right]^2}
	\end{pmatrix}\\
	&\diff(\notset{B}) = \notset{B} \setminus \set{{\Delta\theta}_c \pmod{\pi}} \qquad\text{with ${\Delta\theta}_c = \arctan\left(-\frac{f}{v - y_c}\right)$}\\
	&\argmax_{\Delta\theta \in \diff(\notset{B})}
	\Bigl|\nabla_{\Delta\theta} J\left(\Delta\theta\right)\Bigr| \in
	\set{{\Delta\theta}_{\min}, ~{\Delta\theta}_{\max}, ~\arctan\left(\frac{v - y_c}{f}\right) \pmod{\pi}}
\end{align}

\subsection{Yaw perturbation \texorpdfstring{$\bm{\Delta\psi}$}{dpsi}}
\begin{figure}[ht!]
	\includegraphics[width=\linewidth]{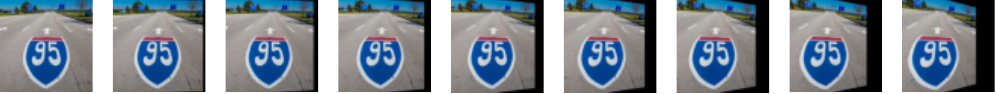}
	\caption{Samples for a yaw perturbation, with $\vect{\kappa}=\Delta\psi \in \notset{B} = [0,15]\degree$}
\end{figure}

\begin{align}
	&\matr{H}^{-1}(\Delta\psi)=
		\begin{bmatrix}
			\cos{\left(\Delta\psi \right)} - \frac{x_{c} \sin{\left(\Delta\psi \right)}}{f} & 0 & \frac{\left(f^{2} + x_{c}^{2}\right) \sin{\left(\Delta\psi \right)}}{f}\\
			- \frac{y_{c} \sin{\left(\Delta\psi \right)}}{f} & 1 & \frac{y_{c} \left(f \left(\cos{\left(\Delta\psi \right)} - 1\right) + x_{c} \sin{\left(\Delta\psi \right)}\right)}{f}\\
			- \frac{\sin{\left(\Delta\psi \right)}}{f} & 0 & \cos{\left(\Delta\psi \right)} + \frac{x_{c} \sin{\left(\Delta\psi \right)}}{f}
		\end{bmatrix} \\
	&\Bigl(u_0(\Delta\psi), v_0(\Delta\psi)\Bigr) =
		\begin{pmatrix}
			x_c + f \cdot \frac{f\sin(\Delta\psi) + (u - x_c)\cos(\Delta\psi)}{f\cos(\Delta\psi) - (u - x_c)\sin(\Delta\psi)},~
			y_c + f \cdot \frac{v - y_c}{f\cos(\Delta\psi) - (u - x_c)\sin(\Delta\psi)}
		\end{pmatrix} \\
	&\tr{\nabla_{\Delta\psi}} T^{-1}(\Delta\psi) =
		\begin{pmatrix}
			\frac{f \left[f^{2} + (u - x_c)^2\right]}{\left[f \cos{\left(\Delta\psi \right)} - (u - x_c) \sin{\left(\Delta\psi \right)}\right]^{2}}
			,&
			\frac{f (v - y_c) \left[f \sin{\left(\Delta\psi \right)} + (u - x_c) \cos{\left(\Delta\psi \right)}\right]}{\left[f \cos{\left(\Delta\psi \right)} - (u - x_c) \sin{\left(\Delta\psi \right)}\right]^{2}}
		\end{pmatrix} \\
	&\diff(\notset{B}) = \notset{B} \setminus \set{{\Delta\psi}_c \pmod{\pi}} \qquad\text{with $\psi_c = \arctan\left(\frac{f}{u - x_c}\right)$}\\
	&\argmax_{\Delta\psi \in \diff(\notset{B})}
	\Bigl|\nabla_{\Delta\psi} J\left(\Delta\psi\right)\Bigr| \in
	\set{\Delta\psi_{\min}, ~\Delta\psi_{\max}, ~\arctan\left(-\frac{u - x_c}{f}\right) \pmod{\pi}}
\end{align}

\subsection{Translation perturbation \texorpdfstring{$\bm{\Delta x}$}{dx}}
\begin{figure}[H]
	\includegraphics[width=\linewidth]{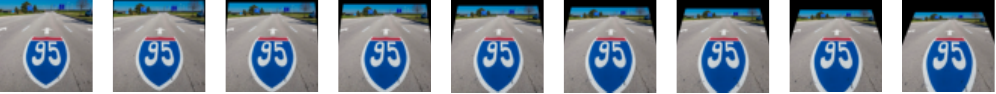}
	\caption{Samples for a $x$-translation perturbation, with $\vect{\kappa}=\Delta x \in \notset{B} = [0,1] m$}
\end{figure}
We notice here that moving the camera forward does not exactly match a simple zooming operation on the image, and is a more involved transform.
\begin{align}
	\matr{H}^{-1}(\Delta x) &=
		\begin{bmatrix}
			1 & - \frac{\Delta x x_{c}}{f z} & \frac{\Delta x x_{c} y_{c}}{f z}\\
			0 & 1 - \frac{\Delta x y_c}{f z} & \frac{\Delta x y_{c}^{2}}{f z}\\
			0 & - \frac{\Delta x}{f z} & 1 + \frac{\Delta x y_c}{f z}
		\end{bmatrix} \quad\text{with $z \neq 0$ (i.e. camera not on ground plane)}\\
	\Bigl(u_0(\Delta x), v_0(\Delta x)\Bigr) &=
		\begin{pmatrix}
			\frac{\Delta x (v - y_c) x_c - f z u}{\Delta x (v - y_c) - f z},&
			\frac{\Delta x (v - y_c) y_c - f z v}{\Delta x (v - y_c) - f z}
		\end{pmatrix} \\
	\tr{\nabla_{\Delta x}} T^{-1}(\Delta x) &=
		\begin{pmatrix}
			\frac{f z \left(u - x_{c}\right) \left(v - y_{c}\right)}{\left[\Delta x (v - y_c) - f z\right]^{2}}, &
			\frac{f z \left(v - y_{c}\right)^{2}}{\left[\Delta x (v - y_c) - f z\right]^{2}}
		\end{pmatrix} \\
	\diff(\notset{B}) &= \notset{B} \setminus \set{\frac{f z}{v - y_c}} \quad\text{when  $v \neq y_c$, $\notset{B}$ otherwise }\\
	\argmax_{\Delta x in \diff(\notset{B})}
	\Bigl|\nabla_{\Delta x} J\left(\Delta x\right)\Bigr| &\in
	\set{{\Delta x}_{\min},{\Delta x}_{\max}}
\end{align}

\subsection{Translation perturbation \texorpdfstring{$\bm{\Delta y}$}{dy}}
\begin{figure}[ht!]
	\includegraphics[width=\linewidth]{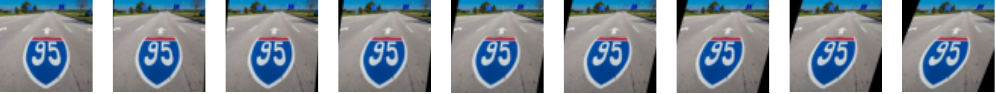}
	\caption{Samples for a $y$-translation perturbation, with $\vect{\kappa}=\Delta y \in \notset{B} = [0,1] m$}
\end{figure}

Since there is no orientation change between both viewpoints, we retrieve a simple shear along the camera $x$-axis: once again an affine transform.
\begin{align}
	\matr{H}^{-1}(\Delta y)&=
		\begin{bmatrix}
			1 & - \frac{\Delta y}{z} & \frac{\Delta y y_{c}}{z}\\
			0 & 1 & 0\\
			0 & 0 & 1
		\end{bmatrix} \quad\text{with $z \neq 0$ (i.e. camera not on ground plane)}\\
	\Bigl(u_0(\Delta y), v_0(\Delta y)\Bigr) &=
		\begin{pmatrix}
			u + \Delta y \left(\frac{y_c}{f} - \frac{v}{z}\right), & v
		\end{pmatrix} \\
	\tr{\nabla_{\Delta y}} T^{-1}(\Delta y) &=
		\begin{pmatrix}
			\frac{y_c}{f} - \frac{v}{z}, & 0
		\end{pmatrix} \\
	\diff(\notset{B}) &= \notset{B}
\end{align}
Any $\Delta y \in \notset{B}$ is suitable to evaluate the Lipschitz constant $L = \sup_{\Delta y \in \diff(\notset{B})}
\Bigl|\nabla_{\Delta y} J\left(\Delta y\right)\Bigr|$.

\subsection{Translation perturbation \texorpdfstring{$\bm{\Delta z}$}{dz}}
\begin{figure}[ht!]
	\includegraphics[width=\linewidth]{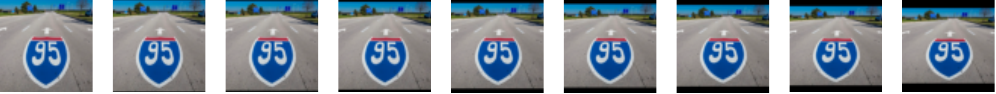}
	\caption{Samples for a $z$-translation perturbation, with $\vect{\kappa}=\Delta z \in \notset{B} = [0,1] m$}
\end{figure}

The simplified form results in a scaling along the camera $y$-axis.
\begin{align}
	\matr{H}^{-1}(\Delta z) &=
	\begin{bmatrix}
		1 & 0 & 0\\
		0 & \frac{z}{\Delta z + z} & \frac{\Delta z y_{c}}{\Delta z + z}\\
		0 & 0 & 1
	\end{bmatrix} \\
	\Bigl(u_0(\Delta z), v_0(\Delta z)\Bigr) &= \begin{pmatrix}
		u,& \frac{z v + \Delta z y_c}{z + \Delta z}
	\end{pmatrix}\\
	\tr{\nabla_{\Delta z}} T^{-1}(\Delta z) &= \begin{pmatrix}
		0,& -\frac{z (v - y_c)}{(z + \Delta z)^2}
	\end{pmatrix}\\
	\diff(\notset{B}) &= \notset{B} \setminus \set{-z} \qquad\text{(i.e. camera not on ground plane)} \\
	\argmax_{\Delta z \in \diff(\notset{B})}
	\Bigl|\nabla_{\Delta z} J\left(\Delta z\right)\Bigr| &\in
	\set{{\Delta z}_{\min}, {\Delta z}_{\max}}
\end{align}

\section{Full closed-form expressions}
\label{supp:closed-forms}
We detail here the remaining closed-form expressions needed in \cref{eq:homography-vbl}. Note that the roll, pitch, yaw angles are not defined in the camera frame $(C)$ ($x$-right, $y$-down, $z$-forward), but in a body frame $(B)$ with $x$-forward, $y$-right, $z$-down. The transform $\matr{T}^{CAM}_{BODY}$ is thus needed to obtain the correct transforms. There is no translation between both frames:
\begin{equation}
\matr{T}^{CAM}_{BODY} =
	\begin{bmatrix}
		0 & 1 & 0 & 0 \\
		0 & 0 & 1 & 0 \\
		1 & 0 & 0 & 0 \\
		0 & 0 & 0 & 1
	\end{bmatrix} = \tr{\left(\matr{T}^{BODY}_{CAM}\right)}
\end{equation}
The transform from first to second viewpoint is given by $\matr{T}^C_{C_0} = [\matr{R}^C_{C_0}|\vect{t}^C_{C \to C_0}]$ with:
\begin{align}
	&\matr{R}^C_{C_0}
		= \matr{R}^{CAM}_{BODY} \cdot \matr{R}^B_{B_0} \cdot \matr{R}^{BODY}_{CAM}
		= \matr{R}^{CAM}_{BODY} \cdot \tr{\left(\matr{R}^{B_0}_{B}\right)} \cdot \matr{R}^{BODY}_{CAM} \\
	&\matr{R}^{B_0}_{B} =
			\begin{bmatrix}
				\cos(\Delta\psi) & - \sin(\Delta\psi) & 0 \\
				\sin(\Delta\psi) & \cos(\Delta\psi) & 0 \\
				0 & 0 & 1
			\end{bmatrix}
			\begin{bmatrix}
				\cos(\Delta\theta) & 0 & \sin(\Delta\theta) \\
				0 & 1 & 0 \\
				- \sin(\Delta\theta) & 0 & \cos(\Delta\theta)
			\end{bmatrix}
			\begin{bmatrix}
				1 & 0 & 0 \\
				0 & \cos(\Delta\phi) & - \sin(\Delta\phi) \\
				0 & \sin(\Delta\phi) & \cos(\Delta\phi)
			\end{bmatrix} \\
	&\vect{t}^C_{C \to C_0} = \matr{T}^{CAM}_{BODY} \vect{t}^B_{B \to B_0} \qquad\text{with $\vect{t}^B_{B \to B_0} = \tr{\left[\Delta x, \Delta y, \Delta z\right]}$}
\end{align}
Similarly, the transform from ground plane to first viewpoint is given by $\matr{T}^W_{C_0} = [\matr{R}^W_{C_0}|\vect{t}^W_{W \to C_0}]$ with:
\begin{align}
	&\matr{R}^W_{C_0} = \matr{R}^{W}_{B_0} \cdot \matr{W}^{BODY}_{CAM} \\
	&\matr{R}^{W}_{B_0} = \begin{bmatrix}
		\cos(\psi) & - \sin(\psi) & 0 \\
		\sin(\psi) & \cos(\psi) & 0 \\
		0 & 0 & 1
	\end{bmatrix}
	\begin{bmatrix}
		\cos(\theta) & 0 & \sin(\theta) \\
		0 & 1 & 0 \\
		- \sin(\theta) & 0 & \cos(\theta)
	\end{bmatrix}
	\begin{bmatrix}
		1 & 0 & 0 \\
		0 & \cos(\phi) & - \sin(\phi) \\
		0 & \sin(\phi) & \cos(\phi)
	\end{bmatrix}
\end{align}

\section{Example: yaw deviation \texorpdfstring{$\bm{\Delta\psi}$}{dpsi}}
\label{supp:yaw-deviation}

For the yaw deviation example, we set the pose of the first camera (i.e. $\phi,\theta,\psi,x,y,z$) to be arbitrary, and set $\Delta\phi=\Delta\theta=\Delta x = \Delta y = \Delta z = 0$ so that only the yaw deviation $\Delta\psi$ explains the motion from first to second viewpoint. In this context, $\vect{t} = \vect{0}$ and \cref{eq:homography-vbl} reduces to the special case $\matr{H}(\Delta\psi) = \matr{K} \cdot \matr{R} \cdot \matr{K}^{-1}$, valid even if scene points are not co-planar (see \cite[Sec. 8.4.5]{hartley2003multiple}). The yaw rotation then corresponds to a rotation about the camera $y$-axis:
\begin{equation}
	\matr{H}(\Delta\psi)
	= \matr{K} \cdot \matr{R^{C}_{C_0}} \cdot \matr{K}^{-1}
	= \matr{K} \cdot \matr{R}_y(-\Delta\psi) \cdot \matr{K}^{-1} \\
\end{equation}

As we need the inverse mapping, we can compute it directly as:
\begin{align}
	\matr{H}^{-1}(\Delta\psi)
	&= \matr{K} \cdot \matr{R}^{-1}_y(-\Delta\psi) \cdot \matr{K}^{-1} \\
	&= \matr{K} \cdot \matr{R}_y(\Delta\psi) \cdot \matr{K}^{-1} \\
	&=
		\begin{bmatrix}f & s & x_c\\0 & f & y_c\\0 & 0 & 1\end{bmatrix}
		\begin{bmatrix}
				\cos{\left(\Delta\psi \right)} & 0 & \sin{\left(\Delta\psi \right)}\\
				0 & 1 & 0\\
				-\sin{\left(\Delta\psi \right)} & 0 & \cos{\left(\Delta\psi \right)}\end{bmatrix}
		\begin{bmatrix}\frac{1}{f} & 0 & - \frac{x_c}{f}\\0 & \frac{1}{f} & - \frac{y_c}{f}\\0 & 0 & 1\end{bmatrix} \\
	&=
		\begin{bmatrix}
			\cos{\left(\Delta\psi \right)} - \frac{x_{c} \sin{\left(\Delta\psi \right)}}{f} & 0 & \frac{\left(f^{2} + x_{c}^{2}\right) \sin{\left(\Delta\psi \right)}}{f}\\
			- \frac{y_{c} \sin{\left(\Delta\psi \right)}}{f} & 1 & \frac{y_{c} \left(f \left(\cos{\left(\Delta\psi \right)} - 1\right) + x_{c} \sin{\left(\Delta\psi \right)}\right)}{f}\\
			- \frac{\sin{\left(\Delta\psi \right)}}{f} & 0 & \cos{\left(\Delta\psi \right)} + \frac{x_{c} \sin{\left(\Delta\psi \right)}}{f}
		\end{bmatrix}
\end{align}

\section{Maximization of $\abs{\nabla_{\kappa} J}$}
\label{supp:gradient-maximization}

In~\cite{batten2024verification}, an upper bound on the Lipschitz constant is estimated as $\sup_{\vect{\kappa} \in \diff\left(\notset{B}\right)} \abs{\tr{\nabla}_{\vect{\kappa}} J\left(\vect{\kappa}\right) \cdot e_m}$ for coordinate $m$. We justify here the maximization of the independent components for the case where there is only one transform parameter $(m=1)$:
\begin{align}
	\abs{\nabla_{\vect{\kappa}} J(\vect{\kappa})} &\triangleq \abs{\nabla_{\vect{\kappa}}\left[\LB(\vect{\kappa}) - G(\vect{\kappa})\right]} \\
	&\triangleq \abs{\nabla_{\vect{\kappa}}\left[\max_{j \in \llbracket 1,q \rrbracket} \set{\tr{\vect{\underline{w}}}_j \vect{\kappa} + \vect{\underline{b}}_j} - G(\vect{\kappa})\right]} \\
	&\leq \biggl|\nabla_{\vect{\kappa}}\max_{j \in \llbracket 1,q \rrbracket} \set{\tr{\vect{\underline{w}}}_j \vect{\kappa} + \vect{\underline{b}}_j}\biggr|
		+ \biggl|\nabla_{\vect{\kappa}} G(\vect{\kappa})\biggr| & \text{by triangle inequality of $\abs{\cdot}$} \\
	&\triangleq \abs{\vect{\underline{w}^\ast}} + \abs{\nabla_{\vect{\kappa}} \left(I \circ T^{-1}\right)(\vect{\kappa})} & \text{with $\vect{\underline{w}^\ast} = \max_{j \in \llbracket 1,q \rrbracket} \abs{\vect{\underline{w}}_j}$} \\
	&\triangleq \abs{\vect{\underline{w}^\ast}} + \abs{\nabla_{\vect{\kappa}} I\Bigl(u_0(\vect{\kappa}), v_0(\vect{\kappa})\Bigr)} \\
	&= \abs{\vect{\underline{w}^\ast}} + \abs{
			\frac{\partial I}{\partial u} \cdot \frac{\partial u_0}{\partial \vect{\kappa}}(\vect{\kappa}) +
			\frac{\partial I}{\partial v} \cdot \frac{\partial v_0}{\partial \vect{\kappa}}(\vect{\kappa})} & \text{by chain rule} \\
	&\leq \abs{\vect{\underline{w}^\ast}} +
		\abs{\frac{\partial I}{\partial u} \cdot \frac{\partial u_0}{\partial \vect{\kappa}}(\vect{\kappa})} +
		\abs{\frac{\partial I}{\partial v}\cdot \frac{\partial v_0}{\partial \vect{\kappa}}(\vect{\kappa})} & \text{by triangle inequality of $\abs{\cdot}$} \\
	&= \abs{\vect{\underline{w}^\ast}} +
		\abs{\frac{\partial I}{\partial u}} \cdot \abs{\frac{\partial u_0}{\partial \vect{\kappa}}(\vect{\kappa})} +
		\abs{\frac{\partial I}{\partial v}} \cdot \abs{\frac{\partial v_0}{\partial \vect{\kappa}}(\vect{\kappa})}
\end{align}

In particular, a majorant can be chosen by maximizing each positive component independently:
\begin{equation}
	\sup_{\vect{\kappa} \in \diff\left(\notset{B}\right)} \abs{\nabla_{\vect{\kappa}} J(\vect{\kappa})}
	\leq
	\abs{\vect{\underline{w}^\ast}} +
	\sup_{\vect{\kappa} \in \diff\left(\notset{B}\right)}
	\abs{\frac{\partial I}{\partial u}}
	\cdot
	\sup_{\vect{\kappa} \in \diff\left(\notset{B}\right)}
	\abs{\frac{\partial u_0}{\partial \vect{\kappa}}(\vect{\kappa})} +
	\sup_{\vect{\kappa} \in \diff\left(\notset{B}\right)}
	\abs{\frac{\partial I}{\partial v}}
	\cdot
	\sup_{\vect{\kappa} \in \diff\left(\notset{B}\right)}
	\abs{\frac{\partial v_0}{\partial \vect{\kappa}}(\vect{\kappa})}
\end{equation}

\section{Upper bound locations for gradient $\abs{\nabla_{\Delta\psi} T^{-1}}$}
\label{supp:gradient-derivation}
In the yaw perturbation example, we follow \cref{supp:gradient-maximization} and derive a majorant for the gradient of the inverse transform $\abs{\tr{\nabla_{\Delta\psi}} T^{-1}}$, more particularly along each coordinate by maximizing $\abs{\frac{\partial u_0}{\partial\Delta\psi}}$ and $\abs{\frac{\partial v_0}{\partial\Delta\psi}}$ independently.\\

\textbf{$\mathbf{u}$-coordinate.} We study the variations of the gradient $\frac{\partial u_0}{\partial\Delta\psi}$ given by \cref{eq:grad_T}. We compute its derivative $\forall \Delta\psi \in \diff(\notset{B})$ and study its sign:
\begin{align}
	\frac{\partial^2 u_0}{\partial {\Delta\psi}^2}
	&= 2 f \left[f^2 + (u - x_c)^2\right] \frac{f\sin(\Delta\psi) + (u - x_c)\cos(\Delta\psi)}{\left[f\cos(\Delta\psi) - (u - x_c)\sin(\Delta\psi)\right]^3} \\
	&= \underbrace{\frac{2 f \left[f^2 + (u - x_c)^2\right]}{\left[f\cos(\Delta\psi) - (u - x_c)\sin(\Delta\psi)\right]^2}}_{> 0} \left[\frac{f\sin(\Delta\psi) + (u - x_c)\cos(\Delta\psi)}{f\cos(\Delta\psi) - (u - x_c)\sin(\Delta\psi)}\right]
\end{align}
The derivative cancels out if $f \sin(\Delta\psi) + (u - x_c)\cos(\Delta\psi) = 0$, which yields:
\begin{equation}
	\Delta\psi = {\Delta\psi}_0 \equiv \arctan\left(\frac{x_c - u}{f}\right) \pmod{\pi}
\end{equation}
For our scope of robotics applications and cameras without wide fields of view, we limit ourselves to $\Delta\psi \in \left[-\frac{\pi}{2}, \frac{\pi}{2}\right]$, otherwise the plane of interest is not in the field of view.

In other cases where $\frac{\partial^2 u_0}{\partial {\Delta\psi}^2} \neq 0$, the gradient $\abs{\frac{\partial u_0}{\partial\Delta\psi}}$ is monotonic, and its maximum occurs at the interval boundary $\psi_{\min}$ or $\psi_{\max}$. Overall, there are three candidate locations for the maximum of $\sup_{\Delta\psi \in \diff(\notset{B})} \abs{\frac{\partial u_0}{\partial\Delta\psi}}$: $\psi_{\min}$, $\psi_{\max}$, or ${\Delta\psi}_0$ if ${\Delta\psi}_0 \in \left[\psi_{\min}, \psi_{\max}\right]$.\\

\textbf{$\mathbf{v}$-coordinate.} We study the variations of the gradient $\frac{\partial v_0}{\partial\Delta\psi}$ given by \cref{eq:grad_T} and compute its derivative $\forall \Delta\psi \in \diff(\notset{B})$:
\begin{align}
	\frac{\partial^2 v_0}{\partial {\Delta\psi}^2}
	&= f (v - y_c) \cdot
		\frac{2 \left[f \sin(\Delta\psi) + (u - x_c) \cos(\Delta\psi)\right]^2 + \left[f \cos(\Delta\psi) - (u - x_c) \sin(\Delta\psi)\right]^2}
		{\left[f \cos(\Delta\psi) - (u - x_c) \sin(\Delta\psi)\right]^{3}}
\end{align}
The derivative cancels out if:
\begin{align}
	&2 \left[f \sin(\Delta\psi) + (u - x_c) \cos(\Delta\psi)\right]^2 + \left[f \cos(\Delta\psi) - (u - x_c) \sin(\Delta\psi)\right]^2 = 0 \\
	\Rightarrow~& \text{(and)}\begin{cases}
		f \sin(\Delta\psi) + (u - x_c) \cos(\Delta\psi) = 0 \\
		f \cos(\Delta\psi) - (u - x_c) \sin(\Delta\psi) = 0
	\end{cases} \\
	\Rightarrow~& \begin{cases}
		f \sin(\Delta\psi) = -(u - x_c) \cos(\Delta\psi) \\
		f \cos(\Delta\psi) = (u - x_c) \sin(\Delta\psi)
	\end{cases} \\
	\Rightarrow~& \begin{cases}
		\tan(\Delta\psi) = \frac{-(u - x_c)}{f} \\
		\tan(\Delta\psi) = \frac{f}{u - x_c}
	\end{cases} \\
	\Rightarrow~&\frac{-(u - x_c)}{f} = \frac{f}{u - x_c} \\
	\Rightarrow~& f^2 = -(u - x_c)^2 \\
	\Rightarrow~& u = x_c \text{ and } f = 0 \text{ (impossible as $f > 0$)}
\end{align}

Since $\forall \Delta\psi \in \diff(\notset{B}),~\frac{\partial^2 v_0}{\partial {\Delta\psi}^2} \neq 0$, the gradient $\abs{\frac{\partial v_0}{\partial\Delta\psi}}$ is monotonic, and its maximum occurs at the interval boundary $\psi_{\min}$ or $\psi_{\max}$.

\section{Derivation of \texorpdfstring{$\bm{f_\text{bound}}$}{f bound}}
By considering the subdomain midpoint, Equation 18 in \citet{batten2024verification} proposes to use $f_{\text{bound}} = f\left(\frac{\kappa_1 + \kappa_2}{2}\right) + \frac{L}{2}\left(\kappa_2 - \kappa_1\right)$. In our implementation, we propose a slightly different bound by working rather with the subdomain extremities. Let $\kappa \in \left[\kappa_1, \kappa_2\right] \subset \RR$. Using the Lipschitz property of $f$, we derive the following bound for $f(\kappa)$:
\begin{equation}
	\begin{aligned}
		& \begin{cases}
			f(\kappa) \leq f(\kappa_1) + L \cdot \abs{\kappa_1 - \kappa} = f(\kappa_1) + L \cdot \left(\kappa - \kappa_1\right)\\
			f(\kappa) \leq f(\kappa_2) + L \cdot \abs{\kappa_2 - \kappa} = f(\kappa_2) + L \cdot \left(\kappa_2 - \kappa\right)
		\end{cases}\\
		\Rightarrow~ & 2 \cdot f(\kappa) \leq f(\kappa_1) + f(\kappa_2) + L \cdot \left(\kappa_2 - \kappa_1\right) & \text{summing up both lines}\\
		\Rightarrow~ & f(\kappa) \leq \frac{1}{2} \cdot \Bigl[f(\kappa_1) + f(\kappa_2) + L \cdot \left(\kappa_2 - \kappa_1\right)\Bigr] \triangleq f_\text{bound}
	\end{aligned}
\end{equation}

\section{Models}
\label{supp:models}
To verify robustness against the VNN datasets \cite{brix2023vnncomp}, we use the official benchmark networks available from the competition repositories at \url{https://github.com/VNN-COMP} (see \cref{tab:models}). There is no existing VNN benchmark for runway visibility, so we train a custom runway classifier on the LARD dataset \cite{ducoffe2023lard}, network and training details are given below.

\begin{table}[ht!]
	\centering
	\small
	\begin{tabular}{rl}
		\toprule
		\textbf{Dataset} & \textbf{Model} \\
		\midrule
		\textbf{MNIST} & \texttt{mnist-net\_256x2.onnx} \\
			& \texttt{mnist-net\_256x4.onnx} \\
			& \texttt{mnist-net\_256x6.onnx} \\
		\addlinespace
		\textbf{CIFAR-10} & \texttt{cifar\_base\_kw.onnx} \\
			& \texttt{cifar\_wide\_kw.onnx} \\
			& \texttt{cifar\_deep\_kw.onnx} \\
		\addlinespace
		\textbf{GTSRB} & \texttt{3\_30\_30\_QConv\_16\_3\_QConv\_32\_2\_Dense\_43\_ep\_30.onnx} \\
		\addlinespace
		\textbf{LARD} & Custom (based on \texttt{resnet\_4b.onnx}) \\
		\bottomrule
	\end{tabular}
	\caption{\label{tab:models} Detail of models used in our experiments.}
\end{table}

\textbf{Runway classifier details.} We only use data for which along-track distance to the runway is available (e.g. \texttt{LARD\_train.csv} and \texttt{LARD\_test\_synth.csv}). We stratify the data per approach, and keep 80\% of approaches for training and validation, and 20\% for testing. In order to make models compatible with the current scalability of formal verification techniques, we scale all images down to $32 \times 32$. We add new labels for classification instead of object detection: positive instances are images with a runway occupying more than 25 pixels, negative instances are images with a runway smaller than 1 pixel. Runway sizes in-between are ignored for better class separation. The model is a small ResNet architecture \texttt{resnet\_4b.onnx} introduced in the 2021 VNN competition. We train for 100 epochs with an Adam optimizer and an initial learning rate of $3 \times 10^{-4}$. A scheduler reduces this rate by a factor of 0.5 every 20 epochs without loss improvement. We do not use any robustification techniques during training, but we use data augmentation (horizontal flip, random rotation up to 10\degree, random translation between 0.1 and 0.2, and random perspective up to 0.2 distortion). Despite the dataset imbalance, we use a simple weighted cross-entropy loss function. This approach produces reasonable results with a 0.96 area under the ROC curve (161 true positives, 36 false positives, 433 true negatives, 26 false negatives). The verification instances are then sampled from the true positives and the true negatives.

\section{Extended results}
\label{supp:extended}
To complete \cref{tab:overall}, this section presents extended robustness results on additional networks of the VNN-COMP benchmarks. We indicate the number of timeouts in brackets, if any. A high number of timeouts indicates a potential need to increase the timeout limit and leave more time for the verifier to output a certified result.

\begin{table}[ht!]
	\centering
	\small
	\setlength{\aboverulesep}{0pt}
	\setlength{\belowrulesep}{0pt}
	\renewcommand{\theadfont}{\small\bfseries}

	\begin{tabular}{ll lll lll c}
		\toprule
		\multirow{2}{*}{\textbf{Perturbation}} &
        \multirow{2}{*}{\textbf{Range} $\mathcal{B}$}
		& \multicolumn{3}{c}{\thead{MNIST}}
		& \multicolumn{3}{c}{\thead{CIFAR-10}}
		& \thead{GTSRB} \\

        \cmidrule(lr){3-5} \cmidrule(lr){6-8} \cmidrule(lr){9-9}

        &

        & \texttt{\textbf{x2}}
        & \texttt{\textbf{x4}}
        & \texttt{\textbf{x6}}

        & \texttt{\textbf{base}}
        & \texttt{\textbf{wide}}
        & \texttt{\textbf{deep}}

        & \texttt{\textbf{30\_30}} \\
		\midrule

		$\bm{\Delta \phi}$ (roll)    & $\bm{[0, 5]}^\circ$
        & 61 & 63 & 48 & 69 & 73 (11) & 57 (4) & 0 \\

		$\bm{\Delta \theta}$ (pitch) & $\bm{[0, 5]}^\circ$
        & 5 & 4 & 2 & 10 (1) & 5 (47) & 2 (4) & 0 \\

		$\bm{\Delta \psi}$ (yaw)     & $\bm{[0, 5]}^\circ$
        & 25 & 15 & 7 (3) & 7 (1) & 6 (50) & 1 (4) & 0 \\

        \addlinespace

		$\bm{\Delta x}$              & $\bm{[0,1]~m}$
        & 50 & 52 & 40 & 21 & 19 (33) & 9 (13) & 0 \\

		$\bm{\Delta y}$              & $\bm{[0,1]~m}$
        & 73 & 74 & 62 & 83 & 85 (3) & 71 (4) & 0 \\

		$\bm{\Delta z}$              & $\bm{[0,1]~m}$
        & 53 & 53 & 42 & 24 & 33 (28) & 13 (18) & 0 \\

		\bottomrule
	\end{tabular}
	\caption{\label{tab:extended}\textbf{Ratio (\%) of robust cases to non-affine perturbations.}}
\end{table}

\section*{Errata and updates}
\label{supp:errata}
\begin{itemize}
	\item \textit{04/05/2026:} equation \ref{eq:roll_u0_v0} of $(u_0,v_0)$ for roll was missing the $x_c$ and $y_c$ offsets, and also used $(v - x_c) \cos(\Delta\phi)$ instead of $(v - y_c)\cos(\Delta\phi)$ for $v_0$. The corrected expression is:
	\[
	\Bigl(u_0(\Delta\phi), v_0(\Delta\phi)\Bigr) = \begin{pmatrix}
		x_c + (u - x_c) \cos(\Delta\phi) - (v - y_c) \sin(\Delta\phi),~\\
		y_c + (u - x_c) \sin(\Delta\phi) + (v - y_c) \cos(\Delta\phi)
	\end{pmatrix}\\
	\]
	\item \textit{04/23/2026:} additional corrections to the PWL code were implemented, added a missing $\abs{w^*}$ term to $L_m$, added a missing $P(i,j)$ term in bilinear interpolation.
\end{itemize}

\end{document}